\ifx\XeTeXrevision\undefined\pdfoutput=1\fi
\documentclass[11pt,onecolumn]{cleantechnicalreport}
\usepackage[authoryear,round]{natbib}
\usepackage{hybridfrontmatter}

\usepackage{setspace}
\usepackage{bm}
\usepackage{multirow}
\usepackage{array}
\usepackage{tabularx}
\usepackage{float}
\usepackage{enumitem}
\usepackage{comment}
\usepackage{framed}
\usepackage{tikz}
\usetikzlibrary{arrows.meta, positioning, shapes.geometric, calc, fit, backgrounds}
\definecolor{promptbg}{gray}{0.955}
\newcommand{\SingleSpacedXI}{\singlespacing}
\newenvironment{promptshade}{\colorlet{shadecolor}{promptbg}\begin{snugshade}\SingleSpacedXI\footnotesize}{\end{snugshade}}
\newcommand{\promptname}[1]{\noindent{\normalfont\sffamily\bfseries\small #1}\par\vspace{3pt}\hrule height 0.4pt\vspace{1pt}}
\title{When AI Analysts Disagree: Agentic Multiverse Analysis and the Robustness of Empirical Research}
\author[1]{Ruijiang Gao}
\author[1]{Steven Chong Xiao}
\affil[1]{Naveen Jindal School of Management, The University of Texas at Dallas}
\reportwebsite{https://github.com/ruijiang81/agentic_multiverse_analysis}
\reportcontact{ruijiang.gao@utdallas.edu, steven.xiao@utdallas.edu}

\begin{abstract}
Every empirical analysis on observational data rests on dozens of defensible analytical choices (``forks'') that its readers cannot see. We develop an agentic multiverse analysis (AMA) framework that makes a conclusion's sensitivity to those choices visible at scale, and we position it as a decision-support instrument for human experts. A pilot study first deploys 150 autonomous AI coding agents on one dataset and one set of research questions and finds that they disagree through the same defensible methodological choices human analysts make, though over a narrower range. AI peer review leaves their dispersion almost unchanged. Exposure to top-rated exemplar reports then collapses it, largely through imitation rather than analysis. Built on those lessons, AMA has agents enumerate a published paper's possible fork space, score every fork against a validity rubric, execute the validated specification grid, and cross-examine one another's work under strict separation of duties. Applied to ten empirical studies from a leading management journal's replication packages (1,755 specification estimates), the framework reproduces every paper's original estimates, and shows that some conclusions hold across every specification the assessment validates, while others are sensitive to individual analytical choices that the published texts do not discuss. The case studies suggest our AMA-based measures can serve as robustness proxies for empirical findings. AI makes multiverse analysis practical, where it once required resources few research teams can afford: our exercises take hours of agent time and tens of dollars per paper. We discuss the implications for replication policy and for AI-assisted peer review in information systems.
\end{abstract}

\begin{document}

\maketitle

\section{Introduction}
\label{sec:intro}

When many independent human analysts test the same hypothesis on the same data, they disagree. Twenty-nine teams given one soccer dataset disagreed on magnitude and significance \citep{silberzahn2018many}; 73 teams testing one social-policy hypothesis disagreed on the sign itself \citep{breznau2022observing}; and when 164 finance research teams tested six market-quality hypotheses on shared data, the dispersion across teams rivaled the statistical uncertainty within any single analysis, the quantity \citet{menkveld2024nonstandard} name \emph{nonstandard errors} (NSE). The dispersion is not special to any specific field and arises from the analytical degrees of freedom every empirical study leaves open, a concern as old as \citet{leamer1983lets} that has been documented across disciplines \citep{simmons2011false,brodeur2020methods}. The direct remedy is to evaluate the defensible alternatives instead of one path through them: multiverse and specification-curve analysis \citep{steegen2016increasing,simonsohn2020specification,young2017model,patel2015assessment}. However, this remedy is expensive and beyond the reach of most research teams. For example, \citet{menkveld2024nonstandard} use 343 researchers and 27 years equivalent of workload to perform the analysis. Generative AI agents can now perform empirical research: they read data documentation, write and debug estimation code, choose measures and specifications, and draft the report \citep{lu2024aiscientist,ape2026}. This paper studies whether agentic AI can help solve the nonstandard-error problem. 

AI could take two possible paths to solving the nonstandard-error problem. The first is that AI analysts converge on the single ``best'' specification, since they are trained to pick the best response, and dispersion across analysts disappears. The second is the opposite: AI agents enumerate every defensible analytical choice, execute all of them, and read the result as a map of the analysis space and a measure of a conclusion's empirical robustness.

While AI agents are trained to maximize user preferences, for many empirical questions no uniquely best specification exists. For example, ``does this platform strategy increase fairness'' can be answered by various fairness metrics and identification strategies. In practice, we also do not want this to happen as it would replace a visible disagreement among analysts with an invisible sensitivity to whichever option the models happen to share. Nevertheless, we test this possibility using AI coding agents in a pilot study. Following the spirit of \citet{menkveld2024nonstandard}, we deploy 150 autonomous Claude Code agents to independently test six hypotheses about market quality on one shared dataset, NYSE Trade and Quote (TAQ) millisecond data for the SPDR S\&P 500 exchange-traded fund (ticker SPY, 2015--2024). The agents pass through three stages: independent analysis, written AI peer review, and exposure to the five top-rated reports. The full experiment cost \$1,558 and about 25 wall-clock hours, a small fraction of what the original human study required.

The agents disagree much as the human teams do. Dispersion in the estimates concentrates in a small number of methodological choices: a single dominant fork can account for nearly all of the cross-agent variance, for most hypotheses conditioning on that choice sharply narrows the spread, and agents that make the same methodological choice differ little in how they execute it.\footnote{Throughout, a \emph{fork} is a discrete analytical choice with two or more defensible options (e.g., which significance test, which estimand, which clustering method); a \emph{specification} is one complete path through the forks, and the \emph{fork space} is the set of all defensible paths, the ``garden of forking paths'' mentioned in \citet{steegen2016increasing}.} Model families (e.g., Opus vs.\ Sonnet) also carry stable preferences, each favoring particular measures, so much of the dispersion comes from cross-model instead of within-model variation. AI agents also explore a narrower set of forks than humans, leaving a portion of the estimation paradigms that human teams use unexplored. AI peer review leaves the dispersion almost unchanged. Exposure to the top-rated reports collapses the dispersion substantially, though not completely, as multiple defensible choices still survive. Overall, AI does not eliminate nonstandard errors by automatically settling on a single best path.

Inspired by the pilot, we design \emph{agentic multiverse analysis} (AMA). One high-capability agent enumerates the fork space for a given paper; an independent agent validates every fork against the specification-curve validity criteria of \citet{simonsohn2020specification}, discarding invalid arms and designating the statistically appropriate inference before any estimate exists; an implementation agent executes the surviving grid deterministically; and a fourth agent reviews the implementation, with a blocking loop that must clear before any result is read. The output of AMA is a measure of empirical robustness: for each empirical conclusion, the share of valid specifications that keep its sign, the share under which it survives at the designated inference, and the identity of the forks that may invalidate the empirical findings. Eight specialized agent sessions per paper replace the pilot's 450 agent-runs per dataset, because the forks are enumerated once rather than rediscovered by every agent. 

We validate AMA on ten empirical studies drawn from the replication packages of a leading management journal, spanning information systems (IS), finance, and applied economics. To avoid reputational harm to the original authors, we refer to the studies as Papers A--J and describe their settings generically. Every assessment anchors on a verified reproduction provided by the authors (original estimates recomputed to the precision each package permits). Three general findings hold across the corpus (Table~\ref{tab:core} collects the specifics). Robustness is heterogeneous across papers. The survival of the published conclusion in the valid specification universe runs from unanimous to zero, with most papers in between. For published papers, signs mostly hold in AMA, and significance often does not hold. The choices with the greatest power to overturn a conclusion are largely unmentioned in the published text, and in five of the ten papers an explicit robustness assurance in the published text is not borne out in the paper's own replication data. At hours of agent time and tens of dollars per paper, the method measures the empirical robustness of published hypotheses cheaply enough for potential routine use.

This paper makes four contributions. First, we provide large-scale evidence that agentic AI analysts produce structurally diverse econometric analyses, and we characterize that diversity (disagreement concentrates on a few nameable forks; model families have stable ``empirical styles''). Second, we propose a reusable multi-agent framework (fixed prompts, fixed rubric, artifact-based gates) that leverages the diversity to measure the robustness of empirical findings, in only hours of agent time and tens of dollars per paper. Third, we demonstrate the framework as a robustness assessment for IS and finance research, showing it can strengthen good papers as well as flag potential issues. Fourth, we draw implications for AI-assisted peer review and for journals' replication-package policies.

\begin{table}[!htbp]
\centering
\caption{The pilot and the ten assessed papers at a glance. Sign is the share of valid specifications carrying the published direction; survival additionally requires $p<0.05$ under the inference designated before any estimate exists (both defined in Section~\ref{sec:metric}, per-paper detail in Table~\ref{tab:overview}). Robustness differs widely across the ten papers, and in every case the specification that overturns a conclusion was itself rated valid by the independent fork validation. In three papers the option rated invalid is the paper's own headline choice. The decisive fork is the one the assessment identifies in its report; when it is an inference designation or a diagnostic applied to every cell it is not a grid dimension, so it can differ from the grid fork with the largest invalidation rate in Table~\ref{tab:overview}. Appendix~\ref{app:decomp} decomposes the failure axes. Design abbreviations: RD, regression discontinuity; DiD, difference-in-differences; IV, instrumental variables.}
\label{tab:core}
\footnotesize
\resizebox{\textwidth}{!}{
\begin{tabular}{lp{4.6cm}ccp{6.4cm}}
\toprule
Study & Field and design & Sign (\%) & Survive (\%) & Decisive fork (fork validator's rating) \\
\midrule
Pilot & 150 AI agents $\times$ 6 hypotheses; millisecond market data; measures analyst disagreement (interquartile range up to 10.7\%/yr) rather than a published claim & --- & --- & --- \\
\midrule
A & IS: RD in time & 100 & 30 & Control set; rated \emph{invalid} for headline use (fails the design's own validity diagnostics) \\
B & IS: DiD + RD & 72 & 9 & Assignment-level clustering; the assessment-designated admissible inference \\
C & IS: matched DiD & 95 & 3 & Assignment-level clustering; the paper's own finer-level clustering rated \emph{invalid} \\
D & IS: panel DiD & 99 / 40 & 78 / 7 & The paper's own analysis subsample; rated \emph{invalid} (selection on a post-treatment variable) \\
E & IS: staggered DiD & 22 & 0 & Excluding already-treated comparisons; the rubric-preferred robust estimators \\
F & Fin: panel + shock & 100 / 68 & 67 / 20 & Treatment functional form in the secondary leg; one variant is not estimable in the shipped data \\
G & Fin: triple-difference & 95 & 17 & Sample-cleaning rule and few-cluster inference; both assessment-designated \\
H & Econ: randomized assignment + IV & 77 & 0 & Randomization inference over the actual assignment; the assessment-designated inference \\
I & IS: staggered DiD & 64 & 10 & Group-by-time trend control; the assessment-designated confound repair \\
J & IS: matched panel & 93 & 47 & Pre-outcome placebo; an assessment-mandated diagnostic (surfaced by the implementation review) \\
\bottomrule
\end{tabular}
}
\end{table}

Section~\ref{sec:related} positions the work; Sections~\ref{sec:pilot} and \ref{sec:framework} present the pilot and the framework; Section~\ref{sec:audits} reports the assessments; Sections~\ref{sec:implications} and \ref{sec:conclusion} discuss implications and conclude. Appendix~\ref{app:genai} documents the use of generative AI: models, prompts, and human oversight.

\section{Related Work}
\label{sec:related}

\paragraph{Many analysts, the multiverse, and replication.}
The concern that empirical conclusions depend on unreported analytical choices is old \citep{leamer1983lets}; its modern formulations are multiverse analysis \citep{steegen2016increasing}, specification-curve analysis \citep{simonsohn2020specification}, multimodel analysis \citep{young2017model}, and the ``vibration of effects'' \citep{patel2015assessment}, with \citet{simmons2011false} and \citet{brodeur2020methods} documenting why such transparency is needed. Crowdsourced many-analyst studies replace enumeration with people \citep{silberzahn2018many,breznau2022observing,huntingtonklein2021hidden}; \citet{menkveld2024nonstandard} formalize the resulting dispersion as nonstandard errors with 164 finance research teams. Such many-analyst studies are costly and consume person-years of volunteer labor (roughly 27 person-years for \citealp{menkveld2024nonstandard}). In AMA, agents supply the labor, and the enumeration is done by an agent that did not write the paper, then disciplined by a validity rubric rather than averaged over good and bad specifications, the failure mode discussed in \citet{simonsohn2020specification}. Since \citet{dewald1986replication}, replication in economics has found published results hard to reproduce from authors' own materials, although a large reproducibility audit at a management journal recovered the reported results in most articles once data and code were disclosed \citep{fisar2024reproducibility}, which is the pattern our own reproduction step also finds. Meanwhile econometric methods are rapidly evolving (e.g., clustering levels \citep{abadie2023when}, few-cluster corrections \citep{cameron2008bootstrap}, randomization inference \citep{young2019channeling}, and staggered-adoption estimators \citep{goodman2021difference,callaway2021difference}). AMA also allows for a revisit of old studies with modern analysis methods. 

\paragraph{Generative AI as a research instrument and concurrent systems.}
A fast-growing literature deploys large language model (LLM) based systems as researchers: end-to-end paper generation \citep{lu2024aiscientist}, autonomous policy evaluation \citep{ape2026}, and simulated economic agents \citep{horton2023homo}. A parallel literature measures what these systems do rather than what they produce, benchmarking their capabilities against human ones on canonical experimental tasks \citep{wang2026unraveling} and auditing the reasoning behind a delegated decision to see which stated values predict the action taken \citep{leng2026suva}. Our pilot is an instance of that measurement problem posed on an analytical task, and the framework that follows is what we build once the measurement shows that more agents do not produce agreement. The behavioral literature on LLM sycophancy \citep{sharma2024towards} predicts the imitation failure we observe when agents see exemplar papers, and motivates the framework's separation of duties. Several concurrent works study related ideas. \citet{bertran2026many} deploy populations of LLM analysts on fixed datasets, with an AI screen for methodological validity, and document dispersion traceable to identifiable analytic choices. Their multiverse is sampled from analyst runs, which is close to our pilot run. Our approach is to enumerate a candidate fork space once, then screen it for validity ex ante against the \citet{simonsohn2020specification} criteria. \citet{huang2026ai} run 158 single-shot GPT instances on the \citet{menkveld2024nonstandard} task metadata and find concentration on a narrow set of paths. The result is consistent with our findings in Stage 1 for the pilot run, despite a different implementation. Similar findings in the pilot run have been concurrently identified in various domains outside information systems \citep{cui2026same, bertran2026many, rabanser2026towards,allen2026intelligence}, with the same broad conclusion validated by different datasets and AI model classes. A parallel stream of work automates reproduction, recovering published results from methods sections and raw data \citep{kohler2026read,holtdirk2026automated}, repairing broken replication pipelines \citep{shah2026automating}, and performing perturbation-based sanity checks \citep{rewolinski2026sanity}. To our knowledge, AMA is the first multi-agent system that enumerates a published paper's fork space once and screens it for validity before any estimate exists, and we are not aware of a published IS journal article that deploys LLM agents as research analysts.

\paragraph{Research credibility and generative AI in IS scholarship.}
Inside IS the credibility conversation is active: a dedicated replication outlet with an explicit manifesto \citep{dennis2015manifesto}, a debate over what replication can establish \citep{olbrich2017merits,brendel2023immodest}, and the IS Replication Project's result that the field's findings held up better than the crisis narrative predicted \citep{dennis2020replication}, a direct precedent for our finding that assessments can strengthen sound papers. MIS Quarterly has articulated a pluralistic research-transparency policy \citep{burtonjones2021transparency}, and INFORMS journals have moved unevenly toward mandatory code-and-data disclosure \citep{loch2025transparency}. Methodologically, AMA extends the IS tradition of computationally intensive research instruments \citep{berente2019computationally} to the evaluation of research itself. IS editors and methodologists are working out what generative AI should do in the scholarly pipeline such as responsible-use charters \citep{susarla2023janus,gopal2025inventing,shmueli2023ijds}, design pathways for AI artifacts \citep{abbasi2024pathways}, and AI's role in peer review, from feasibility studies \citep{drori2024human,checco2021ai,liang2024llmfeedback} to governance \citep{kankanhalli2024peer} and democratization arguments \citep{sarker2024democratizing}. Our contribution to that conversation is a multi-agent framework and empirical evidence on how agentic multiverse analysis reveals empirical findings' robustness. The framework also speaks to the IS literature on delegation between people and agentic artifacts \citep{baird2021delegation,fugener2022cognitive}, which asks which parts of a task should be handed to a machine and what the person retains. AMA gives that question a concrete answer in one setting. The agents enumerate, execute, and cross-examine; the humans set scope, review the configuration, enforce the gates, and decide what a sensitivity means. Our own oversight record suggests the division of labor is uneven in a way worth theorizing, because the substantive errors were surfaced by the machine cross-examination layer before the human review of the same artifacts. 
\section{Pilot Study: Do AI Analysts Eliminate Nonstandard Errors?}
\label{sec:pilot}

The pilot replicates the \citet{menkveld2024nonstandard} many-analyst design with AI agents to test the first possibility from the introduction, and its role in the paper is to establish three observations. First, AI analysts also exhibit nonstandard errors as human analysts do, and do not converge to one ``optimal'' solution. Second, the forks behind those errors are largely the ones human researchers adopt in practice. Third, AI peer review with the objective of getting the best paper does not eliminate the nonstandard errors; the convergence that does occur appears to come from imitation of exemplars rather than from evaluation. We summarize the design and the empirical evidence that support these observations. Appendix~\ref{app:pilot} reports protocols, costs, and abridged instructions. Concurrent work on different datasets and AI model harnesses/classes \citep{cui2026same,rabanser2026towards,bertran2026many,allen2026intelligence} also corroborates the findings in the pilot. 

\subsection{Design}

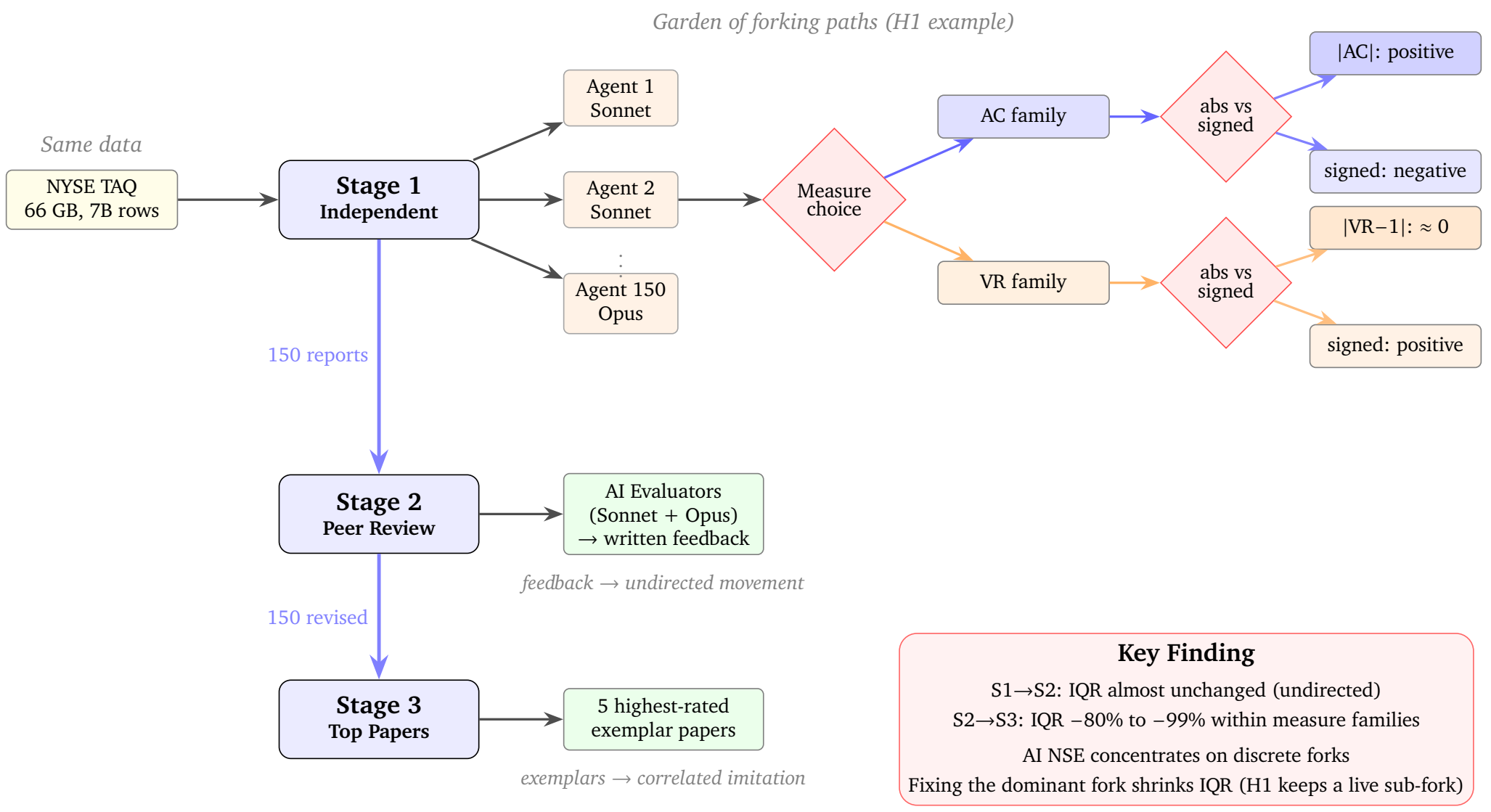
\begin{figure}[!htbp]
\centering
\resizebox{.82\textwidth}{!}{
\begin{tikzpicture}[scale=0.85, every node/.style={scale=0.85},
    >=Stealth,
    node distance=1.2cm and 1.8cm,
    stage/.style={rectangle, rounded corners=4pt, draw=black, fill=blue!8,
                  minimum width=2.8cm, minimum height=1.1cm, align=center, font=\small\bfseries},
    agent/.style={rectangle, rounded corners=2pt, draw=gray!70, fill=orange!10,
                  minimum width=1.6cm, minimum height=0.6cm, align=center, font=\scriptsize},
    feedback/.style={rectangle, rounded corners=2pt, draw=black!50, fill=green!8,
                     minimum width=2.8cm, minimum height=0.8cm, align=center, font=\scriptsize},
    result/.style={rectangle, rounded corners=2pt, draw=black!50, fill=yellow!10,
                   minimum width=2.4cm, minimum height=0.6cm, align=center, font=\scriptsize},
    fork/.style={diamond, draw=red!70, fill=red!8, minimum width=1.2cm,
                 minimum height=0.8cm, align=center, font=\scriptsize},
    arr/.style={->, thick, color=black!70},
    darr/.style={->, dashed, color=black!40},
]

\node[stage] (s1) {Stage 1\\[-2pt]\scriptsize Independent};

\node[agent, above right=0.4cm and 1.0cm of s1] (a1) {Agent 1\\Sonnet};
\node[agent, right=1.0cm of s1] (a2) {Agent 2\\Sonnet};
\node[agent, below right=0.4cm and 1.0cm of s1] (a3) {Agent 150\\Opus};
\node[font=\scriptsize, color=gray] at ($(a2)!0.5!(a3) + (0, -0.1)$) {$\vdots$};

\node[result, left=1.2cm of s1] (data) {NYSE TAQ\\66 GB, 7B rows};
\draw[arr] (data) -- (s1);

\draw[arr] (s1) -- (a1);
\draw[arr] (s1) -- (a2);
\draw[arr] (s1) -- (a3);

\node[fork, right=1.0cm of a2] (f1) {\scriptsize Measure\\[-2pt]\scriptsize choice};
\draw[arr] (a2) -- (f1);

\node[result, above right=0.3cm and 0.8cm of f1, fill=blue!12] (r1a) {AC family};
\node[result, below right=0.3cm and 0.8cm of f1, fill=orange!12] (r1b) {VR family};
\draw[arr, color=blue!60] (f1) -- (r1a);
\draw[arr, color=orange!60] (f1) -- (r1b);

\node[fork, right=0.6cm of r1a] (f2) {\scriptsize abs vs\\[-2pt]\scriptsize signed};
\draw[arr, color=blue!60] (r1a) -- (f2);

\node[result, above right=0.1cm and 0.6cm of f2, fill=blue!18] (r2a) {\scriptsize $|$AC$|$: positive};
\node[result, below right=0.cm and 0.6cm of f2, fill=blue!10] (r2b) {\scriptsize signed: negative};
\draw[arr, color=blue!50] (f2) -- (r2a);
\draw[arr, color=blue!50] (f2) -- (r2b);

\node[fork, right=0.6cm of r1b] (f3) {\scriptsize abs vs\\[-2pt]\scriptsize signed};
\draw[arr, color=orange!60] (r1b) -- (f3);

\node[result, above right=0.cm and 0.6cm of f3, fill=orange!18] (r3a) {\scriptsize $|$VR$-1|$: $\approx 0$};
\node[result, below right=0.1cm and 0.6cm of f3, fill=orange!10] (r3b) {\scriptsize signed: positive};
\draw[arr, color=orange!50] (f3) -- (r3a);
\draw[arr, color=orange!50] (f3) -- (r3b);

\node[stage, below=2.8cm of s1] (s2) {Stage 2\\[-2pt]\scriptsize Peer Review};
\node[feedback, right=1.0cm of s2] (fb) {AI Evaluators\\(Sonnet + Opus)\\$\to$ written feedback};
\draw[arr] (s2) -- (fb);
\node[font=\scriptsize\itshape, color=gray, below=0.12cm of fb] {feedback $\to$ undirected movement};

\draw[arr, very thick, color=blue!50] (s1.south) -- node[left, font=\scriptsize] {150 reports} (s2.north);

\node[stage, below=1.5cm of s2] (s3) {Stage 3\\[-2pt]\scriptsize Top Papers};
\node[feedback, right=1.0cm of s3] (tp) {5 highest-rated\\exemplar papers};
\draw[arr] (s3) -- (tp);
\node[font=\scriptsize\itshape, color=gray, below=0.12cm of tp] {exemplars $\to$ correlated imitation};

\draw[arr, very thick, color=blue!50] (s2.south) -- node[left, font=\scriptsize] {150 revised} (s3.north);

\node[rectangle, rounded corners=6pt, draw=red!60, fill=red!5,
      minimum width=5cm, align=center, font=\small,
      right=5.0cm of s3] (insight) {
  \textbf{Key Finding}\\[2pt]
  \scriptsize S1$\to$S2: IQR almost unchanged (undirected)\\
  \scriptsize S2$\to$S3: IQR $-$80\% to $-$99\% within measure families\\[2pt]
  \scriptsize AI NSE concentrates on discrete forks\\
  \scriptsize Fixing the dominant fork shrinks IQR (H1 keeps a live sub-fork)
};

\node[font=\footnotesize\itshape, color=gray, above=0.1cm of data] {Same data};
\node[font=\footnotesize\itshape, color=gray, above=1cm of f1] {Garden of forking paths (H1 example)};

\end{tikzpicture}
}
\caption{Pilot design: 150 AI agents (100 Sonnet, 50 Opus) analyze the same NYSE TAQ data independently (Stage 1), receive written AI peer review (Stage 2), and see the five highest-rated exemplar papers (Stage 3). Peer review leaves dispersion almost unchanged; exemplar exposure collapses it through imitation. IQR denotes the interquartile range of estimates across agents. The tree shows H1's garden of forking paths \citep{steegen2016increasing}: the measure fork (autocorrelation, AC, vs.\ variance ratio, VR) and its absolute-vs.-signed sub-forks flip the sign of the median estimate (magnitudes in Figure~\ref{fig:pilot_spec}).}
\label{fig:pilot_overview}
\end{figure}

We deploy 150 autonomous Claude Code agents (100 on the Sonnet~4.6 model, 50 on Opus~4.6) against NYSE TAQ millisecond trade and quote data for SPY, 2015--2024 (available by subscription through Wharton Research Data Services, WRDS): every trade and national-best-bid-and-offer quote update over 2,516 trading days, roughly 66~GB and 7 billion rows, mounted read-only inside isolated Singularity containers. The isolation is important. In trial runs without it, agents located and copied other agents' outputs. Each agent independently tests six market-quality hypotheses (efficiency, quoted spread, realized spread, volume, volatility, price impact) and delivers a report, code, and a structured estimates file. The three-stage protocol follows established human multiverse analysis in a top business journal \citep{menkveld2024nonstandard}  (Stage~1 independent analysis, Stage~2 revision after written AI peer review, Stage~3 revision after seeing the five top-rated reports). Figure~\ref{fig:pilot_overview} summarizes the design and, for the market-efficiency hypothesis, the fork structure that generates the dispersion quantified below.

\subsection{Empirical Findings in the Pilot Run}

\begin{table}[!htbp]
\centering
\caption{Stage-1 effect-size dispersion across 150 AI agents (\%/yr, relative to the measure's mean level): one dominant fork per hypothesis explains up to 95\% of cross-agent variance. ``Dominant fork'': largest univariate $R^2$ among fork dummies. IQR is the interquartile range across agents; ``IQR within arms'': range of IQRs conditioning on that fork's arms. $^{\dagger}$Excludes the six-agent Kyle-$\lambda$ arm (IQR 6.98).}
\label{tab:pilot_nse}
\small
\begin{tabular}{lrrcl}
\toprule
Hypothesis & Median & IQR & IQR within arms & Dominant fork ($R^2$) \\
\midrule
H1 Market efficiency & $-$0.06 & 2.43 & 1.45--2.73 & abs/signed $\approx$ measure (0.17, 0.16) \\
H2 Quoted spread & $-$6.21 & 0.43 & 0.003--0.11 & log vs.\ level (0.88) \\
H3 Realized spread & 4.75 & 5.28 & 1.26--3.26 & trade weighting (0.14) \\
H4 Volume & 5.84 & 10.69 & 0.11--0.25 & dollar vs.\ share (0.95) \\
H5 Volatility & 3.08 & 0.54 & 0.005--0.12 & log vs.\ level (0.21) \\
H6 Price impact & $-$10.31 & 10.34 & 1.35--2.59$^{\dagger}$ & measure type (0.62) \\
\bottomrule
\end{tabular}
\end{table}

\begin{figure}[!htbp]
\centering
\includegraphics[width=.8\textwidth]{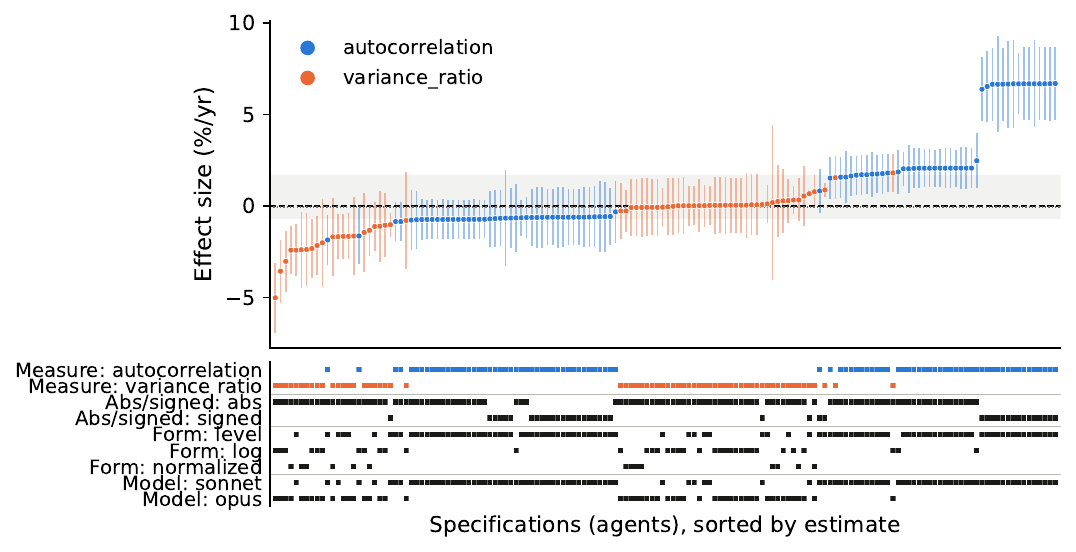}
\caption{Specification curve \citep{simonsohn2020specification} for the market-efficiency hypothesis (H1), Stage 1. Top: the 150 agents' estimates, sorted, with 95\% confidence intervals; the grey band is the pooled interquartile range and the dotted line the median. Bottom: each agent's fork choices. Color marks the measure fork; within the autocorrelation family, the absolute-versus-signed computation separates a positive arm from a negative one (medians $+1.6$ vs.\ $-0.6$\%/yr).}
\label{fig:pilot_spec}
\end{figure}

\paragraph{AI analysts also exhibit nonstandard errors.} 
Stage-1 interquartile ranges (IQRs) run from 0.43 to 10.69\%/yr (Table~\ref{tab:pilot_nse}). The agents choose different forks, each valid in the existing literature. For example, in H1, autocorrelation and variance ratio are valid proxies for market efficiency in the finance literature \citep{menkveld2024nonstandard}. 
Beyond different measurement forks, within H1's autocorrelation measure family, the purely computational choice of absolute versus signed computation moves the median estimate from $+1.6$ to $-0.6$\%/yr, an opposite-sign disagreement on the same data with the construct held fixed (Figure~\ref{fig:pilot_spec} draws the full curve). We also observe wide dispersion when research questions admit multiple valid measures. For volume, dollar-volume agents find $+6.1$\%/yr and share-volume agents $-4.6$\%/yr. Such dispersion persists even within a single measure construct. The three hypotheses whose dominant fork is analytical rather than measure-related (H2, H3, H5) still show IQRs of 0.43--5.28\%/yr, driven by functional form and trade weighting. 
The dispersion is comparable in spirit to the human counterpart that studies similar questions. \citet{menkveld2024nonstandard} report NSEs of the same order as individual standard errors, and across our six hypotheses the cross-agent IQR ranges from 0.6 to 15 times the median agent's standard error, exceeding 3 for four of the six (all three measure-ambiguous hypotheses among them). The forks proposed by AI agents are similar to the choices human researchers make in practice (which volume measure, which functional form, which trade weighting). However, some human-proposed forks never appear among those the AI agents propose. For example, all 150 agents independently choose ordinary least squares (OLS) trend regression as the estimation paradigm, and the relative-change paradigm used by 58\% of human teams never appears. This estimation paradigm also shows a large IQR in the human estimates. Part of this gap may reflect the instruction sheet, which asks for an average per-year change in percentage points and a test against no change, a framing that favors a trend regression. Model families also carry stable ``empirical styles''. For example, 87\% of Sonnet agents choose autocorrelation measures for market efficiency; 100\% of Opus agents choose variance ratios. This model-driven difference is significant for all six hypotheses, so a committee drawn from multiple models may be more diverse than repeated draws from one.

\begin{figure}[!htbp]
\centering
\includegraphics[width=\textwidth]{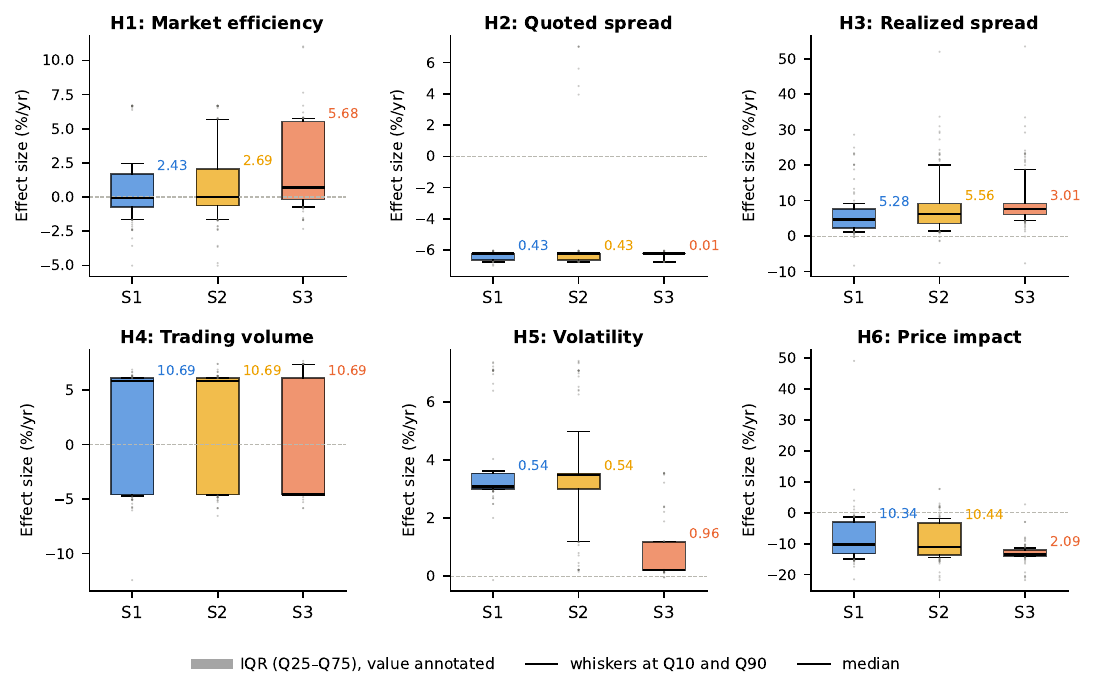}
\caption{Effect-size distributions by stage, one panel per hypothesis. Boxes are the interquartile range with its value annotated, whiskers reach the 10th and 90th percentiles, and the line is the median. Peer review (S2) leaves the dispersion almost unchanged everywhere. Exemplar exposure (S3) moves it in both directions, because the pooled range mixes measure families within which it collapses to.}
\label{fig:pilot_iqr}
\end{figure}

\paragraph{Simple AI review does not reduce diversity.} If AI peer review is effective, the hope is that it can help us identify the ``best'' fork, if one exists, in the garden of forking paths. AI peer review (Stage~2) leaves the IQR almost unchanged on every hypothesis (Figure~\ref{fig:pilot_iqr}). Agents revise their analyses, with 42\% of them changing their measure, but the movements are undirected: on the two hypotheses whose dispersion is generated by a measure-family split the shares moving toward and away from the cross-agent median are balanced (H1 39\% vs.\ 43\%, H4 41\% vs.\ 40\%), and pooled across all six the movement is if anything away from the median (28\% vs.\ 36\%), never toward it. Exemplar exposure (Stage~3) collapses the IQR by 80--99\% within converging measure families, but through imitation: for the market-efficiency hypothesis, 71\% of autocorrelation agents (41\% of all agents) switch to the exemplars' variance-ratio measure; for the volume hypothesis, 87\% of dollar-volume agents switch to share volume while 68\% of share-volume agents switch to dollar volume, a near-random reshuffle with no analytical rationale. Thus, it appears that convergence achieved this way is agreement on whom to copy rather than on what is correct, likely due to the sycophantic behavior of AI agents \citep{sharma2024towards}. One round of AI review did not reduce nonstandard errors in our setting. For system design, this means agents that see one another's conclusions may converge for the wrong reasons. We note this is a direct measurement on state-of-the-art coding agents using simple prompts, and more advanced AI harness design may mitigate this problem in the future. 

\paragraph{A cross-vendor check.} We also check that the Stage-1 dispersion and its fork structure are not an artifact of one vendor's models. We use 35 OpenAI Codex agents (\texttt{gpt-5.2-codex} and \texttt{gpt-5.3-codex}) to run the identical Stage-1 protocol (Appendix~\ref{app:pilot} gives details) as a robustness check. For hypotheses in which the construct is unambiguous, the two vendors agree almost exactly (median quoted-spread trend $-6.2$ vs.\ $-6.2$\%/yr; volatility $+2.9$ vs.\ $+3.1$). Some important forks reappear among Codex agents as well: trade weighting again splits realized-spread medians ($+6.0$ equal- vs.\ $+2.3$ volume-weighted), and the absolute-versus-signed choice again separates H1's autocorrelation medians across zero ($-0.7$ vs.\ $+6.7$\%/yr, four signed agents). We observe some distributional differences in fork choices across models. For example, 34 of 35 Codex agents choose share volume, while among the Claude agents only 60 of 150 do. As a result, the volume IQR among Codex agents falls from 10.7 to 1.7\%/yr; the single Codex agent that uses dollar volume lands at $+5.8$\%/yr, close to the Claude dollar-volume median ($+6.1$). 

\subsection{From Pilot to Agentic Multiverse Analysis Design}

As shown in the pilot study, nonstandard errors do not disappear with AI analysts even though AI systems are trained to produce the best responses. We argue this is a natural consequence, as it is in human many-analyst studies. For most empirical research questions, the choice of empirical measures, analytical tools, and assumptions is subjective and admits many valid interpretations, and no single best option exists. While the pilot study provides a scalable many-analyst exercise compared to the human-analyst alternative, it also generates significant redundancy, with many agents producing duplicate estimates using the same method. In the next section, we propose a design that approaches the diverse nature of empirical research with a scalable framework while addressing the redundancy problem to achieve high efficiency.

\section{Agentic Multiverse Analysis (AMA) Framework}
\label{sec:framework}

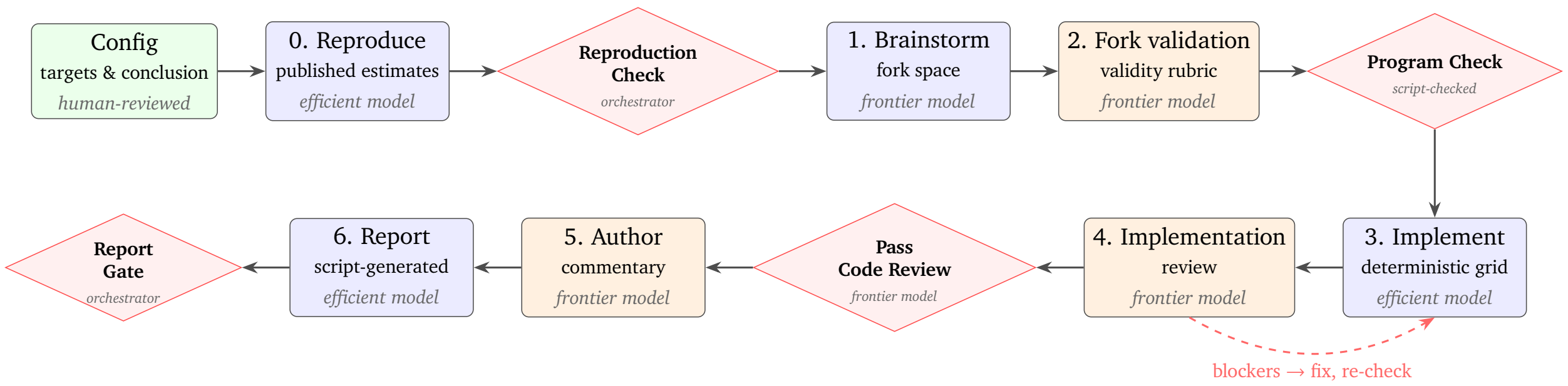
\begin{figure}[!htbp]
\centering
\resizebox{\textwidth}{!}{
\begin{tikzpicture}[
    >=Stealth, node distance=0.6cm and 0.7cm,
    role/.style={rectangle, rounded corners=3pt, draw=black!70, fill=blue!8,
                 minimum width=2.7cm, minimum height=1.35cm, align=center, font=\small},
    audit/.style={rectangle, rounded corners=3pt, draw=black!70, fill=orange!12,
                  minimum width=2.7cm, minimum height=1.35cm, align=center, font=\small},
    gate/.style={diamond, draw=red!70, fill=red!6, aspect=2.2, inner sep=0.5pt,
                 align=center, font=\scriptsize\bfseries},
    arr/.style={->, thick, color=black!70},
]
\node[role, fill=green!8] (sc) {Config\\\scriptsize targets \& conclusion\\[1pt]{\scriptsize\itshape\color{black!60}human-reviewed}};
\node[role, right=of sc] (s0) {0. Reproduce\\\scriptsize published estimates\\[1pt]{\scriptsize\itshape\color{black!60}efficient model}};
\node[gate, right=of s0] (g0) { \\Reproduction\\ Check\\[1pt]\tiny\mdseries\itshape\color{black!60}orchestrator};
\node[role, right=of g0] (s1) {1. Brainstorm\\\scriptsize fork space\\[1pt]{\scriptsize\itshape\color{black!60}frontier model}};
\node[audit, right=of s1] (s2) {2. Fork validation\\\scriptsize validity rubric\\[1pt]{\scriptsize\itshape\color{black!60}frontier model}};
\node[gate, right=of s2] (g2) {\\Program Check\\[1pt]\tiny\mdseries\itshape\color{black!60}script-checked};
\node[role, below=1.3cm of g2] (s3) {3. Implement\\\scriptsize deterministic grid\\[1pt]{\scriptsize\itshape\color{black!60}efficient model}};
\node[audit, left=of s3] (s4) {4. Implementation\\\scriptsize review\\[1pt]{\scriptsize\itshape\color{black!60}frontier model}};
\node[gate, left=of s4] (g4) {\\ Pass\\ Code Review\\[1pt]\tiny\mdseries\itshape\color{black!60}frontier model};
\node[audit, left=of g4] (s5) {5. Author\\\scriptsize commentary\\[1pt]{\scriptsize\itshape\color{black!60}frontier model}};
\node[role, left=of s5] (s6) {6. Report\\\scriptsize script-generated\\[1pt]{\scriptsize\itshape\color{black!60}efficient model}};
\node[gate, left=of s6] (g6) {\\Report \\Gate\\[1pt]\tiny\mdseries\itshape\color{black!60}orchestrator};

\draw[arr] (sc) -- (s0);
\draw[arr] (s0) -- (g0);
\draw[arr] (g0) -- (s1);
\draw[arr] (s1) -- (s2);
\draw[arr] (s2) -- (g2);
\draw[arr] (g2) -- (s3);
\draw[arr] (s3) -- (s4);
\draw[arr] (s4) -- (g4);
\draw[arr, dashed, bend right=30, color=red!60] (s4.south) to node[below, font=\scriptsize, color=red!60]{blockers $\to$ fix, re-check} (s3.south);
\draw[arr] (g4) -- (s5);
\draw[arr] (s5) -- (s6);
\draw[arr] (s6) -- (g6);
\end{tikzpicture}
}
\caption{The AMA pipeline: no agent checks its own work, and the brainstormer never sees the validator's verdicts before enumeration is complete. Blue boxes generate, orange boxes check, diamonds are gates; the green config box is the one artifact a human reviews before the run starts. Each diamond names a gate, and Appendix~\ref{app:gates} states its pass condition. The detailed models used are in the GenAI Use Appendix.}
\label{fig:pipeline}
\end{figure}

This section describes AMA. Section~\ref{sec:audits} applies it to ten published papers. AMA takes as input a published paper's replication package (data, code, and the paper's headline claims) and produces (i) a reproduction of the published estimates, re-implemented where a package's code cannot be run directly, (ii) a grid of validated alternative specifications with estimates and appropriate inference for each, (iii) a conclusion-survival analysis, and (iv) a commentary comparing decisive forks against the paper's own discussions. Figure~\ref{fig:pipeline} shows the agents and gates. Seven in-run agents, a configuration-drafting agent that precedes the run, and four gates. Appendix~\ref{app:prompts} documents the orchestration contract and reproduces the role prompts.

\subsection{Multi-Agent Systems}

AMA runs as a multi-agent system. Each agent has one job and its own input and output. We make sure that no agent checks its own work, and an orchestrator passes artifacts between agents and enforces the gates in Figure~\ref{fig:pipeline}. We describe the agents in sequence.

\paragraph{Step 0: Reproduction agent and reproduction check.} Before the run, a config agent records the paper's headline claims and the reproduction targets. Which estimate counts as the headline is itself a judgment. We anchor on the paper's stated main result, by convention the first column of its headline table, and the configuration file records that choice for human review before the run starts. The reproduction agent then runs, translates, or re-implements the package's analysis. We replicate all the sample studies' main coefficients with deviations at the level of published rounding (see Appendix~\ref{app:gates}). When a package does not report a target directly, the reproduction target is the output of the authors' analysis on the shipped data, and the check compares our re-implementation against that output. Nothing downstream starts until the reproduction check passes. 

\paragraph{Step 1: Brainstorm agent.} The brainstorm agent reads the paper's design, the raw data, and the package code, and enumerates the fork space. Every decision (e.g., sample construction, treatment and outcome definitions, estimator, fixed effects, timing) creates a fork. Every option must be grounded in a named variable or file in the package. The pilot's findings support using one generative agent because a large analyst population mostly rediscovers the same forks in the pilot. A single capable agent with the data in hand can write the candidate list down directly, and what enters the grid is decided by the validation agent rather than by the brainstorm agent itself. Across the ten assessments, the brainstorm agent produced 138--266 options across 10--28 decisions per paper.

\paragraph{Step 2: Validation agent.} An independent validation agent scores every option against a fixed rubric (Appendix~\ref{app:rubric}) that operationalizes the three validity conditions in \citet{simonsohn2020specification}, so that each estimate is theoretically justified, statistically valid, and non-redundant. Options are graded valid, questionable (arguable construct match, retained but flagged), or invalid (excluded, with reasons logged). The validation agent then selects the grid, the cross-product of surviving options that is actually estimated (100--250 cells per identification), and designates the binding inference method before any estimate exists (e.g., assignment-level clustering \citep{abadie2023when}, wild cluster bootstrap under few clusters \citep{cameron2008bootstrap}, randomization inference \citep{young2019channeling}, or heterogeneity-adapted difference-in-differences (DiD) estimators under staggered adoption \citep{goodman2021difference,callaway2021difference}). A fork-coverage gate then checks by script that every brainstormed option carries a verdict, that the grid size is within bounds, and that the inference method is designated in writing (the program check in Figure~\ref{fig:pipeline}).

\paragraph{Step 3: Implementation agent.} The implementation agent writes the fork grid as a deterministic script. Given the validated fork list, the specification set is fixed, and the script writes an estimate, standard error, and designated-inference $p$-value for every cell.

\paragraph{Step 4: Code review gate.} A separate review agent checks the grid script against the validated fork definitions and the raw data, and classifies findings as blockers (results cannot be trusted), major, or minor. A blocker sends the code back to the implementation agent for a fix, and the gate passes only when no blocker remains open. This gate is where the framework's reliability is actually earned, so it is worth reporting what it caught. Across the ten assessments the implementation reviews recorded five blocker-level and around 40 major findings in their severity tallies, every one of them resolved or rebutted in writing before any result was reported. Most were ordinary engineering problems and some of them were substantive. One agent mislabeled an invalid permutation procedure as a randomization test. Another built a placebo test that the true effect contaminated. Both would have produced plausible numbers. The reviewing agent caught both before the human review of the same artifacts.

\paragraph{Steps 5--6: Commentary and report agents and report gate.} The commentary agent searches the published text and the public web for what the authors say about each decisive fork, quotes them verbatim, and verifies every claim against the grid. The report agent then assembles the report, and every number in it must be generated by the verified scripts with no hand-transcription, which is the report gate's pass condition. 

\subsection{Design Principles}

The pilot suggests three design principles for building multi-agent systems in which AI agents evaluate analytical work. Each principle responds to a failure mode we observed in the analyst population and motivates a specific feature of AMA. 

\emph{Elicit the fork space rather than sample it.}
The pilot shows that running more independent analysts does produce diversity, but much of the additional computation is spent rediscovering the same forks. AMA therefore asks a capable agent to enumerate the relevant analytical choices directly and uses a separate agent to determine which of them are defensible. This changes the role of multiple agents. Instead of repeatedly drawing possible analyses from a population, the system makes the underlying choice space explicit and then evaluates it systematically. Future work can also use different sampling from the same model, or different models, to get a more diverse fork space. Section~\ref{sec:audits} reports what a second vendor's independent enumeration finds on the same papers, and some of the significant forks remain the same. 

\emph{Separate generation from evaluation.}
The pilot also shows why simply asking agents to modify their econometric analysis based on the provided review is not enough. Once agents see others' reviews or exemplar reports, their answers can converge through imitation rather than through the resolution of analytical disagreements. AMA therefore only uses agents to judge the validity of the econometric analysis the earlier agent adopts. 

\emph{Verification with inspectable artifacts.}
AMA does not treat an agent's statement that a task has been completed correctly as evidence that it has been completed correctly. Each important transition instead depends on an artifact that another agent or a program can check, like reproduced estimates, validation records, executable grid code, reviewer findings, or script-generated results. When a check fails, the pipeline stops and prevents the next agent from reasoning with an unverified output. The current implementation does not make every gate fully mechanical. Some checks still require an independent agent to compare artifacts or adjudicate whether a problem has been resolved. Therefore, our artifact-based verification is not equivalent to deterministic verification.

\subsection{The Conclusion-Survival Metric}
\label{sec:metric}

For each paper we define the published conclusion as the sign of the headline effect plus its statistical significance, and say a specification \emph{supports} the conclusion if its estimate has the published sign and $p < 0.05$ under the assessment-designated inference. Three summary statistics follow: the \emph{sign-consistency rate} (share of valid specifications with the published sign), the \emph{survival rate} (share supporting the conclusion), and each fork's \emph{invalidation rate} (the spread in the survival rate across that fork's options). A fork with a large spread is one that by itself decides whether the conclusion holds. In one paper, for example, the conclusion survives in 58\% of the specifications that use the published outcome definition and in 2\% of those that use an alternative definition, so that fork's invalidation rate is 56 percentage points, and a reader who wants to believe the conclusion must first believe the outcome-definition choice. This bar is much stricter than asking whether the authors' own specification can be replicated as it asks how much of the defensible analytical universe agrees with them, which is the question a skeptical reader actually has \citep{simonsohn2020specification}. We also want to note some limitations about the metric. First, it is descriptive, so it does not correct for multiplicity across the grid, and it counts each specification once regardless of how closely it resembles its neighbors. And it inherits the fork space, so a specification that no agent proposed casts no vote. 

\subsection{Cost and Time}
\label{sec:cost}

We measure the cost of agent runs with token usage. An assessment uses eight sessions described in Figure~\ref{fig:pipeline}. At list prices for the two models used, the ten assessments cost \$553 in total, from \$28 to \$96 per paper with a median of \$56. Active agent time runs from 2.0 to 4.9 hours per paper with a median of 3.2 hours. Elapsed time from the first session to the last is 2.6 hours at the median and 12.8 hours for the one paper whose implementation review waited on a long grid execution.

\section{Assessing the Robustness of Published Research}
\label{sec:audits}

\subsection{Paper Selection and Anonymization}

We evaluate AMA on ten empirical studies from a leading management journal whose replication packages allow the headline estimate to be reproduced. The ten studies cover information systems, finance, and applied economics, and use a range of research designs (regression discontinuity, staggered and matched difference-in-differences, triple differences, and random assignment with instrumental variables).

The pool is the journal's replication-package archive, which holds more than 400 packages. A typical package contains the analysis data, estimation code, and a README. We required a package with the analysis data and enough code or documentation to reproduce the headline estimate, and a non-experimental design, including natural experiments with quasi-random assignment, with a headline estimate we could locate in a specific table. The ten were not drawn as one random sample, and the composition matters for how the survival rates should be read. Three were selected at the start of the project, while the pipeline was being built. Five were drawn at random from the archive's non-experimental papers in a single year, using a fixed seed. Two were selected deliberately for information systems focus. One further package was set aside during the screening for those two papers because it fell outside the information systems focus; the random draw was not filtered by field. No paper was dropped after its assessment produced results. The ten packages vary widely in size, and some ship richer raw data than the analysis uses. For one paper we rebuilt the outcome from public source records, which let us examine sample-construction choices the processed data forecloses. Appendix~\ref{app:prompts} reports the range of sample sizes and data footprints across the ten studies and documents the computational environment.

We anonymize the papers because our purpose is to evaluate AMA as a method, not to judge individual authors. A conclusion that is sensitive to a defensible analytical choice is not evidence of misconduct or poor research practice. Throughout the paper, we refer to the studies as Papers A--J and describe their settings in general terms. Exact magnitudes, sample sizes, paper identities, and full assessment reports are kept confidential. All ten papers passed the reproduction check and their reported analyses can be recovered from the replication materials to the precision those materials permit.

We also note that agents can make mistakes, and even a correctly estimated sensitivity does not by itself mean the original paper is flawed. Researchers may reasonably disagree about whether an alternative specification is required or how much weight it deserves. We therefore caution against treating our findings as automated verdicts. Our key contribution is that AMA makes the sensitivities measurable and visible, and the judgment remains with the research community.

\subsection{Overview}

\begin{table}[!htbp]
\centering
\caption{Conclusion survival across the assessed multiverses: ``Survive'' requires the published sign and $p<0.05$ under the designated inference as implemented. Universes are paper-specific and total 1,755 cells.}
\label{tab:overview}
\footnotesize
\resizebox{\textwidth}{!}{
\begin{tabular}{clrrrrl}
\toprule
Paper & Field and design & Specs & Sign (\%) & Survive (\%) & Top fork (pp)$^{i}$ & Assessment characterization \\
\midrule
A & IS: RD in time & 216 & 100 & 30$^{a}$ & 56.2 & Survives; magnitude $\approx$ 1/2 \\
B & IS: DiD + RD & 243 & 72 & 9 & 25.9 & Direction holds via second leg \\
C & IS: matched DiD & 176$^{c}$ & 95 & 3 & 18.8 & Direction robust; fails assignment-level inference \\
D & IS: panel DiD & 192$^{h}$ & 99/40 & 78/7 & 31.2/14.6 & First outcome survives; second does not \\
E & IS: staggered DiD & 108$^{b}$ & 22 & 0 & --- & No assessed cell supports it; sign is minority \\
F & Fin: panel + shock & 96/74$^{d}$ & 100/68 & 67/20 & 66.7/35.1 & Core result survives \\
G & Fin: triple-difference & 110 & 95 & 17$^{e}$ & 34.5 & Direction robust; $p \approx 0.05$ \\
H & Econ: randomized assignment + IV & 180 & 77 & 0 & --- & No cell survives randomization inference (min $p = 0.12$) \\
I & IS: staggered DiD & 168$^{f}$ & 64 & 10 & 35.7 & Shrinks to $\sim$1/4--1/3 of published; n.s. \\
J & IS: matched panel & 192 & 93 & 47$^{g}$ & 92.7 & Survives inference; fails placebo \\
\bottomrule
\end{tabular}
}
\begin{minipage}{\textwidth}
\vspace{4pt}
\footnotesize
$^{a}$All 216 positive; 58 of 191 distinct estimates survive per-cell inference, and the full curve is jointly significant under the designated permutation test ($p = 0.035$). $^{b}$Informative cells only, i.e., cells whose samples observe every treated unit before treatment; 96 are distinct after de-duplication, on which 21\% carry the published sign and 79\% the opposite one. $^{c}$Designated inference per cell (assignment-level CRV1, the standard cluster-robust variance estimator, exact randomization inference, or wild cluster bootstrap); 16 further cells infeasible by design. $^{d}$Two evidence legs, 96 primary-leg cells and 74 shock-leg cells, reported separately and scored on the run's headline block; the run's 96 pre-fix sensitivity re-estimates, 16 shock-leg cells rated unidentified, 6 duplicate-handling diagnostics, and the published anchor are not counted, as in the assessment report. $^{e}$Wild bootstrap on every cell; CRV1 alone gives 27/110. $^{f}$Two-way CRV1; 168 cells, 159 distinct estimates after de-duplicating selector coincidences, with 16 survivors on either base. $^{g}$Unit-level CRV1; 1 of 91 survivors passes the pre-outcome placebo; the within-unit design survives in 1 of 96. $^{h}$First outcome 75/96 and second outcome 7/96 under the designated randomization inference (overall 82/192, 43\%); 104/192 under a trend-recentered null; the randomization distribution uses 50 in-time placebo cutoffs, so the smallest attainable $p$-value is 0.02. $^{i}$Invalidation rate of the paper's most decisive fork: the spread in the survival rate across that fork's options, in percentage points (Section~\ref{sec:metric}), computed by the statistics script on the same cells as the survival column; D by outcome family and F by leg. For E and H no cell survives, so every fork's spread is zero and no single fork can be called decisive. This dimension is not always the fork named in Table~\ref{tab:core}, which may be an inference designation or a diagnostic applied to every cell rather than a crossed dimension.
\end{minipage}
\end{table}

Across the ten assessments, the pipeline brainstormed 1,857 fork options in total, of which the fork validations rated 952 valid, 624 questionable, and 281 invalid. Invalid options are excluded, and questionable ones are retained and flagged. Each grid then crosses the valid options of the four to eight dimensions the fork validation judges most consequential, holds the remaining dimensions fixed at a validated default, and adds a bounded set of flagged options as extra cells. The ten grids together produce 1,755 specification estimates (see the notes to Table~\ref{tab:overview}). Table~\ref{tab:overview} summarizes the main results. Published conclusions range from unanimous support to zero support, and sign consistency is systematically higher than survival. The invalidation rate of the most decisive fork runs from 15 to 93 percentage points across the families of the eight papers with any surviving specification, so in each of them one analytical choice moves survival by at least 15 percentage points. Directions mostly hold, and significance often does not, usually because of one or two inference or sample-construction forks. Appendix~\ref{app:decomp} reports the decomposition, separating sign, magnitude, significance, identification-diagnostic, and estimand failures, with magnitude statistics computed on unit-comparable cells in more detail.

How much of this depends on the model that ran the assessment? A second vendor's model (\texttt{gpt-5.6-sol}) repeated the entire framework, from the fork brainstorm through the report, on all ten papers under a blind protocol, with a licensed Stata installation available to it (we use only Python and R for our main experiment as they are open-source softwares). Appendix~\ref{app:xvendor} reports the protocol and the per-paper comparison. 
Every reproduction anchor passed and every paper cleared the code review gate. At the option level the two enumerations agree on 221 of 334 matched fork options (66\%), and they agree more on the estimation machinery (74\%) than on what is measured and on whom (60\%). Six of the ten bottom-line verdicts agree with ours (B, C, E, F, G, and J). Three (A, D, and H) split at the inference designation rather than at any estimate; on A the sign result agrees and only survival splits. The fourth disagreement, Paper I, comes from the fork space instead: the second enumeration never proposes the trend control our grid imposes in every cell, and it reads 46\% survival where ours reads 10\%. In every case we observed, an omitted fork made the second map more forgiving rather than more severe; omitting an arm under which the conclusion survives would have the opposite effect. Therefore, we recommend to combine fork proposals from several models in practice. Having Stata available does not change main qualitative conclusions, which suggests open-source software largely produce similar robustness conclusions and can potentially reduce the adoption barriers for the broader research community. 

We next examine each assessment in detail, beginning with the paper whose conclusions are strengthened by the AMA framework, followed by the paper where the evidence shifts, and then the papers where the assessment finds more structurally detable problems (Papers C, E, and H). We briefly summarize the remaining assessments.

\subsection{Paper A: What an Assessment Adds to an Empirically Stable Paper}
\label{sec:paperA}

Paper A estimates the effect of a policy change introduced on a single date on a later behavioral outcome, identified by a regression discontinuity (RD) in time \citep{hausman2018regression} around the change date.

\begin{figure}[!htbp]
\centering
\includegraphics[width=\textwidth]{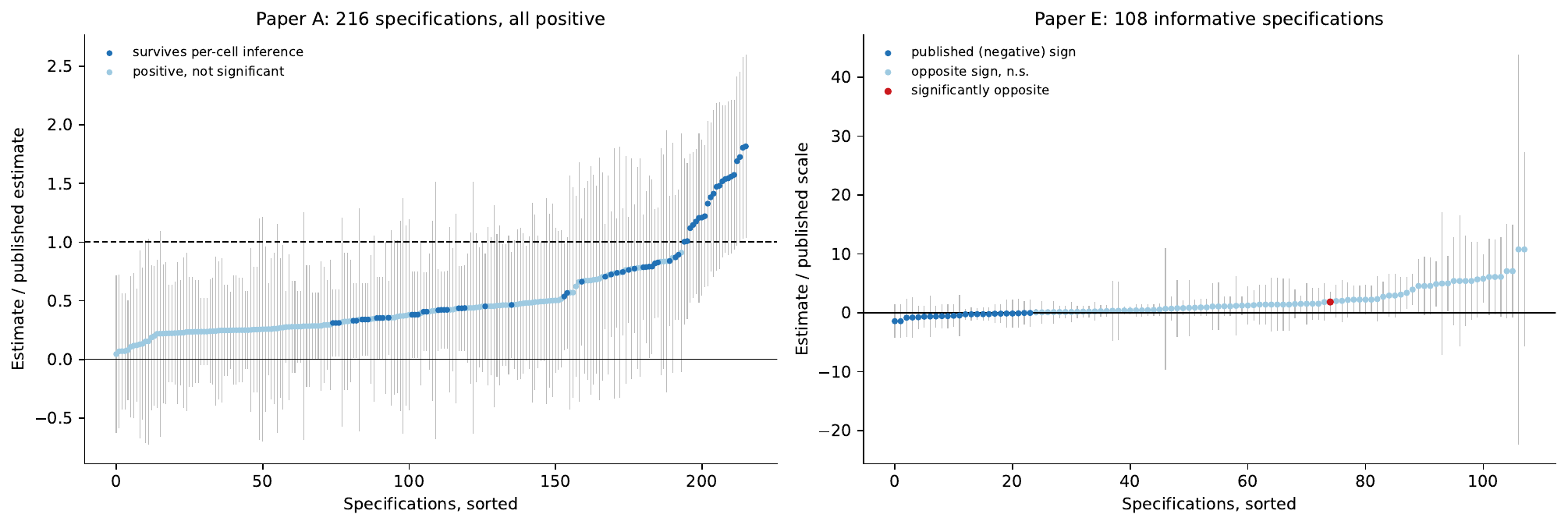}
\caption{Two assessed multiverses, plotted as the ratio of each cell's estimate to the published estimate so that no absolute published quantity appears. Left: Paper A, all 216 grid specifications positive (dark: survives per-cell inference; light: positive but not significant); estimates and intervals are the robust bias-corrected quantities, and the dashed line at 1.0 marks the published estimate. Right: Paper E, the 108 informative specifications, each estimate divided by the absolute published magnitude so that the published direction reads as negative: 24 carry the published sign, 83 are opposite-signed but insignificant, one is significantly opposite (red). Cells with different outcome scales share the axis, so the vertical scale is for display only; Figure~\ref{fig:ec_spec2} restricts to unit-comparable cells. Bars are 95\% confidence intervals under each cell's designated inference.}
\label{fig:speccurves}
\end{figure}

The assessment's most important finding supports the paper. All 216 specifications in the validated grid (every combination of sample definition, outcome definition, bandwidth rule, polynomial order, and kernel; 191 distinct estimates after de-duplicating selector coincidences) produce a positive effect (Figure~\ref{fig:speccurves}, left panel). This is a claim a conventional robustness section has no room to make. 
The specification curve as a whole also passes a test that no single cell can provide. Under the designated permutation inference, which recomputes the estimate at 172 displaced cutoffs, the all-positive curve is jointly significant ($p = 0.035$ on the median estimate, $p = 0.006$ on the share of positive-significant cells). The displacement window was moved farther from the true cutoff at the code review gate, because the true cutoff fell inside most near-offset placebo windows. Under the original window the curve was not jointly significant ($p = 0.09$ and $0.12$). That designation is itself contested. A third model, asked to validate the same fork list under the blind protocol, rated the displaced-cutoff permutation inadmissible on the ground that a calendar cutoff does not supply an exchangeable assignment mechanism, and it declined to authorize headline $p$-values from any method in the menu (Appendix~\ref{app:xvendor}). The sign result does not depend on that choice. The joint-significance result does, and we report it as one defensible ruling rather than a settled one.

The assessment also stress-tests the paper's potential identification problem that the treated sample is defined by an event that could itself respond to the treatment. The placebo checks come back clean: neither the rate of that event nor the size of the treated sample jumps at the cutoff, and a density test at the cutoff is likewise null, so the sample-definition problem that would have invalidated the design does not bite.

The assessment does revise the effect size magnitude, on grounds the paper's own diagnostics support. Several of the paper's control variables are measured after treatment (post-treatment covariates), and two of them fail their own continuity tests at the cutoff ($p = 0.046$ and $0.048$). The fork validation rated the published control set invalid for headline use. Without the controls, the de-duplicated median estimate is roughly half the published one (median ratio 0.50, interquartile range 0.42--0.85). Per-cell significance concentrates at the published outcome definition: 58\% of the distinct cells using it satisfy $p<0.05$, against 2\% under the alternative definition (30\% overall), so a robustness claim the paper makes about this choice does not hold under the alternative. The assessment's summary sentence for Paper A: \emph{the design survives, the direction is unanimous, and the defensible effect size is about half the published one.}

\subsection{Paper B: The Conclusion Holds through the Second Identification Strategy}
\label{sec:paperB}

Paper B estimates the effect of a competitively allocated status on a later time-to-event outcome. The paper uses two identification strategies. The main analysis is a difference-in-differences design comparing units that receive the status with those that do not, estimated as a duration model. A regression discontinuity design at the selection margin provides a second source of evidence.

The AMA results initially look weak. Only 9\% of the validated specifications, or 22 of 243, support the published conclusion. Much of this decline comes from inference rather than a reversal of the estimated effect. The paper clusters standard errors below the level at which treatment is assigned. When we instead cluster at the assignment level \citep{abadie2023when}, the headline specification's $p$-value rises from below 0.05 to about 0.1, under both cluster-robust and bootstrap inference. The main design is also sensitive to other analytical choices. For example, changing the timing definition or the sample construction can reduce the estimated effect to nearly zero.

The overall pattern is more supportive than the survival rate alone suggests. About 72\% of the specification universe retains the published direction. Among the specifications with the opposite sign, none is close to statistically significant, with a minimum $p$-value of 0.20. More importantly, the paper's second identification strategy is much more robust. Across the bandwidths in the authors' own sensitivity analysis, the estimates are large and of the published sign, and the authors' reported intervals exclude the null in nearly every specification. This is a weaker criterion than the 5\% rule used for survival, and the RD leg sits outside the validated grid as corroboration under all three models' designations. 

The assessment therefore changes where the strongest evidence comes from rather than overturning the paper's conclusion. The headline difference-in-differences result is sensitive to the choice of inference and several modeling decisions. The regression discontinuity evidence is more stable across defensible alternatives. For Paper B, AMA therefore shifts the weight of the evidence from the first identification strategy to the second. This distinction is useful because it shows which part of the empirical design provides the more reliable support for the paper's conclusion.

\subsection{Paper C: Significance Fails at the Assignment Level}
\label{sec:paperC}

Paper C uses a staggered difference-in-differences design with a matched control sample, in which treatment is assigned at a coarser geographic level than the unit of observation.

The estimated direction is quite stable across the assessment. About 95\% of the 176 estimable specifications have the published sign. However, only 3\%, or 6 specifications, remain statistically significant under the designated inference. 
Most of the difference between sign consistency and conclusion survival comes from the level of inference. Treatment is assigned at a coarser level than the one at which the paper clusters standard errors. The fork validation therefore rates the paper's clustering as invalid for the headline analysis \citep{abadie2023when} and uses assignment-level inference instead: assignment-level clustered standard errors, exact randomization inference over the possible assignments, or a wild cluster bootstrap, depending on the sample. With this change, the paper's own specification moves from $p<0.001$ under its own clustering to $p = 0.150$, and to 0.305 under randomization inference. Larger $p$-values are expected when clustering moves to the assignment level. 

The point estimates themselves are quite robust. The median estimate across the 176 estimable specifications is about 1.2 times the published estimate. On the unit-comparable cells used in Appendix~\ref{app:decomp} the median ratio is 1.27. Only eight specifications reverse sign, and all of them come from unmatched samples. This suggests that the main issue is not the direction of the estimated relationship, but how much statistical confidence the design can support.

Overall, the assessment leaves the direction of Paper C's result largely intact but weakens the evidence for statistical significance. The main reason is the level at which inference is conducted. 

\subsection{Papers E and H: Minority in the Assessed Universe}
\label{sec:paperE}

Paper E uses a staggered difference-in-differences design in which treatment is assigned at a level above the unit of observation, across a small number of clusters. The published estimate is statistically significant.

Across the 108 informative specifications, 78\% carry the sign opposite to the published one, and none is significant in the published direction under the designated inference (Figure~\ref{fig:speccurves}, right panel). The main reason is the way already-treated units enter the comparison group. Decomposing the published coefficient shows that it comes entirely from comparisons with units that have already received treatment. Once those comparisons are removed, the clean switcher comparisons contribute with the opposite sign. The heterogeneity-robust estimator of \citet{callaway2021difference} shows the same pattern, with opposite-signed estimates in 99\% of its specifications.

The paper's own robustness analysis also raises a concern. It states that the results are similar when a subset of treated clusters is excluded. When agents reproduce that comparison, the estimate falls to essentially zero and loses significance.

Inference provides another source of sensitivity. Under the designated wild cluster bootstrap at the assignment level, the published specification has a $p$-value of 0.15. Because treatment is assigned across only a small number of clusters, no single randomization scheme is identified in these data. Three schemes bracket the published cell's $p$-value between 0.0015, from a restricted scheme that is anti-conservative by construction, and 0.33, from a broad scheme that is conservative by construction, with a stratified middle value of 0.025. The published specification keeps already-treated units as controls and is scored outside the 108 informative cells. Within those cells no estimate is significant in the published direction under the bootstrap or under any of the three schemes, so the zero in Table~\ref{tab:overview} does not depend on the choice among them. Therefore, the paper's own specification produces a minority conclusion in the agents' proposed universes. 

Paper H reaches the same overall outcome for a different reason, and two separate issues arise. The first is the comparison itself. The paper's headline is an interaction estimate, and its fixed effects leave a pre-existing within-cell difference unabsorbed, so part of that difference enters the estimate. Because treatment is randomly assigned, we can measure the difference. We redraw the assignment and re-run the experiment with no effect built in. Under the paper's own specification these placebo runs average about four-fifths of the reported estimate rather than zero. Absorbing the within-cell subgroup difference, by saturating the fixed effects, brings the placebo average back to zero. The estimate then falls by about 40\% and is far from significant. The second issue is the significance test. The published inference uses a wild cluster bootstrap to correct for the small number of clusters.
Because the assignment can be redrawn exactly, the fork validation designates randomization inference \citep{young2019channeling}. Under it none of the 180 valid specifications is statistically significant. The smallest randomization $p$-value is 0.120, and the paper's own specification gives $p = 0.229$.
The same wild cluster bootstrap, which the validation keeps as a required companion statistic, rejects at the 5\% level on the paper's specification. A second model designates that bootstrap as the headline method and reads 38\% survival, and a third model adopts the same designation and rates the exact randomization test questionable (Appendix~\ref{app:xvendor}). The two methods answer different questions. 

Papers E and H therefore both receive relatively few survival in the assessed universes, but for different reasons. Paper E is mainly sensitive to how treated units are used as controls and to the choice of inference. Paper H is more of an inference problem, where the result does not survive a test based on the study's random assignment, under the designation we adopted. Judging the correctness of these forks inevitably requires deep understanding on the problem domain and the assumptions. 

\subsection{The Remaining Assessments in Brief}

\emph{Paper D} estimates the effect of a policy change on two outcomes with a panel difference-in-differences design, and the two produce different results. The first result is robust. Of the 96 specifications for this outcome, 99\% have the published sign and 75 remain significant under the designated randomization inference, which builds its null from placebo cutoffs in the pre-period. Two other models held that no headline $p$-value is informative with very few treated and control units (Appendix~\ref{app:xvendor}), so this designation is also one defensible ruling. The second result is weaker. Only 7 of 96 specifications survive, and the published estimate is not significant in the paper's own full-sample specification. The second outcome is observed only for a selected subset of units, which creates a selection problem; bounds on this conditional effect do not exclude zero. It also raises a concern about a subsample that is central to the paper's interpretation: it conditions on a choice made after treatment and removes nearly all treated post-period observations. The fork validation therefore excludes it from the headline analysis. Overall, the assessment supports the first result but provides much less support for the second result and for the interpretation based on the selected subsample.

\emph{Paper F} provides another case where the main finding holds up well. Its headline result survives in 67\% of the assessed specifications and in every specification using the functional form favored by the authors. The weaker evidence comes from a secondary shock-based analysis, where one specification variant cannot be identified in the shipped data. The assessment therefore leaves the paper's main result largely intact while showing that the secondary evidence is less robust.

\emph{Paper G} uses a triple-difference design with treatment assigned to a small number of entities. Here, 95\% of the specifications retain the published direction, but statistical significance depends heavily on the choice of inference. The published construction is significant with assignment-level CRV1 and under the designated wild cluster bootstrap ($p = 0.033$). The assessment also identifies a sample-construction problem and considers several defensible ways of addressing it. Each correction moves the wild-bootstrap $p$-value just above the 5\% threshold, from 0.051 to 0.098. Under conventional cluster-robust inference the corrected cells remain below 0.05 (0.020 to 0.047), so the marginal verdict reflects the inference method as much as the cleaning rule. A stacked heterogeneity-consistent estimator also finds no significant effect. We therefore view the direction of the result as fairly stable, but the evidence for significance at the 5\% level is much less clear.

\emph{Paper I} estimates the effect of a staggered adoption with a cross-group difference-in-differences design. The reproduction check recovers the published headline result exactly. The assessment, however, finds a pre-existing differential trend between the treated and comparison groups that is close in size to the published effect. The comparison group also changes around the time of treatment. After accounting for these issues, three different identification approaches produce estimates around one-quarter to one-third of the published magnitude, with $p$-values between 0.075 and 0.25. Every cell in the 159-specification universe includes the group-by-time trend control, and the published specification itself is not a grid cell. Across those cells, 64\% retain the published sign and 10\% remain significant under two-way inference. None produces a significant effect in the opposite direction. The evidence therefore still points toward the phenomenon described in the paper, but suggests that the effect is considerably smaller than originally reported. 

\emph{Paper J} estimates the effect of a unit-level action on a later outcome with a reweighted matched panel. Here the assessment diverges from the paper's conclusion for reasons the survival rate alone does not capture. Its published conclusion survives the corrected inference in a substantial part of the multiverse. About 47\% of the 192 valid specifications are significant in the published direction. Looking only at the survival rate would therefore suggest that the result holds. The diagnostic checks tell a different story. One control observation receives nearly all of the control weight in the published construction. After bounding those weights, the estimated magnitude falls to about 70--90\% of the original. More importantly, 90 of the 91 surviving specifications fail a pre-outcome placebo test and produce an apparent effect before treatment. The one comparison that performs well on the placebo is the within-unit design. On the repaired sample, that design produces an estimate close to zero ($p = 0.76$), and only one of its 96 specifications survives. Paper J therefore shows why the survival rate cannot be interpreted on its own. A result can remain statistically significant across a large part of the multiverse while other diagnostics raise serious questions about whether that significance reflects the effect claimed in the paper.

\subsection{What the Authors Say about the Forks that Matter}
\label{sec:commentary}

The final step of the assessment asks what the published paper says about the forks that have the largest effect on its conclusion. Across the ten papers, three patterns stand out.

The first is silence. Many of the choices that matter most for the results receive little or no discussion in the published paper. These include the clustering level, the use of already-treated units as controls, a sample-cleaning rule, the estimator, a selected subsample, and the control pool, each in at least one paper. In each case, changing the choice can substantially alter the assessment of the main result.

The second pattern is that some stated robustness claims do not hold when we reproduce them on the paper's own data. We count a paper when the published text makes an explicit statement about a specific analytical choice and the shipped data do not bear it out. This holds for five papers. Three of them state that a particular analytical choice does not materially affect the results. In one, the text says that clustering at the assignment level does not change the results, but every headline specification loses significance under it. In another, the text says the results are similar without a subset of treated clusters, and excluding them moves the estimate to zero (Section~\ref{sec:paperE}). In the third, a claim about an outcome-definition choice does not hold under the alternative definition. Of the other two, one states that a standard diagnostic of its estimator is benign, and the package's own output contradicts it. In the other, a published diagnostic table does not match the sample produced by the headline procedure, and recomputed with the paper's own formula it is far less favorable.

The third pattern is more encouraging. When authors identify an important analytical choice and explain why they made it, the assessment often supports that choice. Paper A's choice of cutoff is supported by the data. Paper F argues for its preferred functional form, and that is the specification that remains robust throughout the assessment. Paper J discusses a potential bad-control concern, and its reported estimate without those controls matches our recomputation. These cases suggest that the choices authors discuss explicitly are often less problematic than the choices that remain in the background.

Taken together, these findings suggest that the most consequential analytical choices are often not the ones authors spend time defending. They are the choices that receive little attention because they appear routine or are left implicit. One value of the assessment is that it brings those choices into view and shows how much they could matter for the conclusion.

\section{Implications}
\label{sec:implications}

\subsection{For Authors: Assessments as Complements}
\label{sec:authors}

One implication of our results is that a multiverse assessment can be complementary to the paper being assessed. It can also provide stronger evidence for a result that is genuinely robust. Paper A is the clearest example. Its estimated effect remains positive across all 216 specifications, and the assessment finds no evidence that its main identification concern changes at the treatment cutoff. At the same time, the assessment suggests a smaller effect size than the published specification. Paper F shows a similar pattern for its core result. In both cases, the assessment adds information rather than simply trying to overturn the published finding.

This creates a practical use case for authors before submission. A conventional robustness section reports the alternatives that the authors chose to examine. AMA instead asks which other defensible choices are available and shows where the conclusion survives them. For a well-identified paper, this can turn a general statement that ``the results are robust'' into a more transparent description of where they are robust and where they are not. It can also identify weaker secondary claims before those claims become points of disagreement during review. Because an assessment takes only hours of agent time, this type of exercise is potentially feasible for individual research teams.

The same idea applies to older research. Empirical standards change over time. Methods for staggered treatment adoption, clustered inference, and regression discontinuity designs have all evolved, so an analysis that was reasonable when a paper was published may look different under current practice. AMA provides one way to revisit such work without treating methodological change as evidence that the original authors did something wrong. The question is simply whether the conclusion continues to hold when the analysis is evaluated using defensible alternatives available today.

This suggests a more modest version of what we call a \emph{living artifact}. The published paper itself does not change. What can change is the accompanying robustness assessment. As econometric methods and AI systems improve, the same replication package can be reassessed and the resulting robustness map updated. Researchers deciding whether to build on an older finding could then see not only whether the original analysis reproduces, but also whether its conclusion remains stable under current analytical practice.

Our evidence does not establish how often such assessments will strengthen rather than weaken published findings. We observe clear examples of both outcomes in a sample of ten papers. Establishing the calibration of the approach will require a broader benchmark that includes known-robust findings, null results, and deliberately constructed cases with known analytical weaknesses.

\subsection{For Journals and AI-Assisted Review}

A practical starting point would be voluntary pre-submission assessment. Authors could run an AMA-style analysis on their raw data or their own replication package and report the resulting analysis together with the paper. When an alternative specification changes the conclusion, the authors could explain why that alternative is inappropriate for their setting or acknowledge that the conclusion is conditional on a particular analytical choice. Either response is more informative than leaving the choice invisible.

A more ambitious possibility is an editor-initiated assessment conducted alongside conventional referee reports. Such a system could first verify that the headline estimates reproduce and then provide referees with a small set of analytical choices that materially affect the conclusion. The purpose would not be to generate an automated accept-or-reject recommendation. Instead, it would focus human attention on the places where methodological judgment matters most. Referees and authors should still decide whether a particular fork is substantively appropriate.

There are important obstacles before such a system could be used routinely. For confidentiality, unpublished data and manuscripts cannot simply be transmitted to commercial AI systems without appropriate infrastructure and consent. For accountability, agents and their implementations make mistakes, as our own code-review gates demonstrate. Authors would therefore need access to the assessment and an opportunity to challenge its findings. Finally, the assessment itself needs calibration. Our ten-paper sample is enough to demonstrate that the approach can reveal consequential sensitivities, but not enough to estimate its false-alarm rate.

\subsection{Detecting Patterns Consistent with Selective Reporting}
\label{sec:selective}

A further implication concerns selective reporting, although our evidence here needs to be interpreted carefully. AMA makes several features of an empirical analysis visible that are difficult for a reader to observe from the published paper alone: whether the reported specification lies within the broader set of defensible alternatives, whether stated robustness claims hold in the replication data, and, when earlier versions are available, whether consequential analyses changed across versions.

Several cases in our sample illustrate why this information can matter. In one paper, the published sample construction is the only one of several examined variants that crosses the conventional 5\% threshold. Another contains a robustness check that does not test the binding version of its identification concern. In one paper, a test reported in an earlier public version does not appear in the published version and remains null when we reproduce it. Section~\ref{sec:commentary} reports the five papers in which an explicit robustness statement is not borne out by the replication data.

These patterns are informative, but they do not establish selective reporting or author intent. There are many possible explanations for why researchers choose one specification, omit another, or change an analysis across versions. Authors may have strong substantive reasons for a choice that are not fully documented in the paper; analyses may change during review; and space constraints inevitably limit what can be reported.

The contribution of AMA is not to infer intent. It is to make the underlying analytical pattern observable. Editors and readers can see whether a reported estimate is typical of the defensible specification space or depends on a narrow set of choices, and they can check whether robustness statements are supported by the available data. The substantive interpretation of that evidence remains a human judgment.

This also motivates a simple disclosure practice. Alongside the robustness checks they report, authors could document important analytical alternatives they considered but did not use and briefly explain why. Such disclosure would not require authors to estimate every conceivable specification. It would make the consequential forks easier for readers, reviewers, and subsequent researchers to understand.

\paragraph{Governance.}\label{sec:ethics}
Because assessments of published work can affect how individual papers are perceived, they require careful procedural safeguards. Our assessments concern the sensitivity of published conclusions to alternative analytical choices, not the integrity or intentions of the original authors. Before publicly identifying assessed papers or releasing paper-level reports, authors should have an opportunity to inspect the assessment, identify errors, and respond to disputed methodological judgments. The same standard should apply to journal-run assessments. AI-generated robustness evidence should be treated as contestable analytical evidence rather than as an automated verdict.

\subsection{Limitations}
\label{sec:limits}

Our work has limitations. First, the empirical base is small. We assess ten papers from one journal's replication archive, and the ten were not drawn at random from it (Section~\ref{sec:audits}). The results therefore demonstrate what AMA can uncover, but the distribution of survival rates we report should not be read as an estimate of the archive's distribution, and the study does not establish how frequently published research is robust or fragile more generally.

Second, the survival rate is not a formal statistical test and should be read with caution. A low survival rate can reflect genuine specification sensitivity, low statistical power, or both, and the metric does not correct for multiplicity across the specification grid. This is why we report sign consistency alongside statistical-significance survival and examine the forks responsible for changes as complementary tests.

Third, the fork space AMA enumerates depends on the model that proposes it. The pilot (Section~\ref{sec:pilot}) shows that model families carry stable preferences over methods and that agents explore a narrower set of forks than human analysts, leaving whole estimation paradigms unproposed. The assessments point the same way. A second vendor's independent run of the whole framework on the same ten papers reaches our bottom line on six of them and agrees on 221 of 334 matched fork options, and three of its four remaining splits are rulings on which inference method should be the headline rather than disagreements about any estimate, so the fork spaces overlap substantially without being identical. Where a decisive fork is missing from one enumeration, the resulting map was more forgiving rather than more severe in the cases we observed. The inference designation appears to be the least stable layer. In a three-model comparison on four papers, the models agree exactly where the decisive fork has a settled textbook answer, and they split where several defensible inference schemes exist for the same design; the newest model is the most conservative of the three, twice declining to authorize headline $p$-values at all (Appendix~\ref{app:xvendor}). Combining fork proposals from several model families is therefore a useful extension, and IS research has precedent for structured aggregation over a crowd as the remedy for a selection problem of this shape \citep{brynjolfsson2016crowdsquared}.

Finally, AMA can evaluate only what the available data and replication package make observable. It cannot recover unrecorded data-collection decisions, establish the construct validity of the original measures, or determine whether an estimated causal effect exists outside the design encoded in the package. Agents and their code can also be wrong; our own review process found blocker-level implementation errors that had to be corrected before the results could be used. 

For these reasons, an AMA assessment should be interpreted as an attempt to scale up the measurement of empirical robustness. It can identify where a conclusion depends on consequential choices, but deciding what those dependencies mean for the scientific claim remains the responsibility of researchers, reviewers, and editors.

\section{Conclusions}
\label{sec:conclusion}

Empirical conclusions depend on analytical choices that may be opaque to readers, and evaluating all defensible alternatives creates a substantial burden on research teams. This paper asks whether agentic AI can address this problem. We first conduct a pilot study with 150 AI coding agents to show that AI analysts do not converge on a single best specification: like human analysts, they make different choices when a question admits multiple defensible paths.

Inspired by the pilot, we develop AMA to turn this disagreement into a robustness assessment. Agents enumerate and validate possible analytical forks, execute them systematically, and independently check the resulting artifacts. Across ten published studies, some conclusions become stronger, some are difficult to sustain, and others depend on particular choices in measurement, specification, or inference. This is information that traditional reproduction alone cannot provide.

For authors, AMA creates a practical way to stress-test a study before publication and to state more precisely where a conclusion is robust. For journals and reviewers, it suggests a potential role for AI as decision support, as its large-scale assessment can identify the few analytical choices that deserve human attention. This method can also be applied to revisiting older published studies as empirical methods evolve over time, extending a replication package into a robustness map and making an old paper a living artifact.

\paragraph{Data and code availability.} The AMA pipeline is available as an open-source tool at \url{https://github.com/ruijiang81/agentic_multiverse_analysis}. The pilot's agent reports, converted estimates, and analysis notebook are available at \url{https://github.com/ruijiang81/AI_NSE}. The pipeline version used for this paper (fixed prompts, rubric, gate definitions, and orchestration skill) is available from the authors on request. The underlying NYSE TAQ data cannot be redistributed but are available by subscription through WRDS. Assessed papers' replication packages are the property of their authors and the journal. To preserve anonymity, our per-paper configurations, grids, reports, and verification notebooks are not public; like the identity mapping, they remain confidential. 

\paragraph{Appendices.} The appendices contain the pilot study details (Appendix~\ref{app:pilot}), the fork-validity rubric (Appendix~\ref{app:rubric}), gate definitions (Appendix~\ref{app:gates}), the non-robustness decomposition (Appendix~\ref{app:decomp}), specification curves for the assessed papers (Appendix~\ref{app:ecspec}), and implementation details with the role prompts in abridged form (Appendix~\ref{app:prompts}).

\bibliographystyle{plainnat}
\bibliography{references}

\clearpage
\appendix
\counterwithin{table}{section}
\counterwithin{figure}{section}
\renewcommand{\thetable}{\thesection\arabic{table}}
\renewcommand{\thefigure}{\thesection\arabic{figure}}
\section{GenAI Use}
\label{app:genai}

We report the GenAI usage in this project. 

\paragraph{Tools/models used (versions).} Except for the cross-vendor and third-model runs noted below, all agents are instances of Claude Code (Anthropic) running named Claude models at vendor-default sampling; determinism lives in the seeded grid scripts, not in the model. Pilot (March 2026): 100 research agents on Claude Sonnet~4.6 (\texttt{claude-sonnet-4-6}) and 50 on Claude Opus~4.6 (\texttt{claude-opus-4-6}), two peer evaluators (one of each) per report, conversion agents on Opus~4.6. Formalized pipeline, the checking agents (fork brainstorm, fork validation, implementation review, author commentary) on Claude Opus~5 (\texttt{claude-opus-5}), the generation agents (reproduction, grid implementation, report assembly) on Claude Fable~5 (\texttt{claude-fable-5}).  
In Figure~\ref{fig:pipeline}, ``frontier'' marks the roles that ran on Opus~5 and ``efficient'' the roles that ran on Fable~5. 
Cross-vendor replication of the pilot's Stage~1: 36 OpenAI Codex command-line interface (CLI) agents (35 completed) on \texttt{gpt-5.2-codex} and \texttt{gpt-5.3-codex}. 
Cross-vendor replication of the complete framework on all ten papers: OpenAI Codex CLI 0.152.1 on \texttt{gpt-5.6-sol}. 
Third-model verification: \texttt{gpt-6-astra} through Codex CLI 0.153.4 with same protocol (Appendix~\ref{app:xvendor}). 
Appendix~\ref{app:prompts} documents the computational environment in full. Pilot agents ran in Singularity containers from a pinned image on a CPU-only high-performance computing (HPC) cluster managed by the Slurm workload manager, with read-only data mounts and per-agent memory and budget caps. Assessment agents ran in-process on the same cluster under a 16~GB memory cap, with a per-paper copy of the package that the prompts forbid them to modify, and pinned Python and R estimation stacks. Stata license is provided in a robustness check. Structured session logs (model identifier, prompts, tool calls, cost) are retained for the pilot and the formalized runs.

\paragraph{Stages of the research process involved.} Generative AI enters at four stages: the pilot, the ten assessments, the cross-model replications, and manuscript preparation. The pilot is generative-AI research by construction, since its object of study is the analytical behavior of AI agents and its nearest non-AI analog (a 164-team human many-analyst study) consumed substantial volunteer labor across 164 teams. Our replication of that design cost \$1,558 and a few days. 

For the 10-paper analysis, fork spaces of 138--266 options per paper were enumerated, validity-graded, executed, and adversarially re-implemented within hours per paper, a human team may not be able produce any single one of these assessments in a short period of time. Generative AI (Claude Code sessions, mainly on Claude Fable~5 and 5.1 and Claude Opus~5, with short review and verification sub-tasks on Claude Sonnet~5 and Claude Haiku~4.5) assisted in drafting and revising the text, generating figures from the assessment data, and maintaining the statistics script, under the authors' direction and review. 
The authors take full responsibility for all content.
All citations added on AI recommendation were verified against publisher records.

\paragraph{Prompting strategies and agentic workflows.} The agentic workflow is the pipeline of Figure~\ref{fig:pipeline}. All framework prompts are fixed text templates, versioned in the repository and rendered from a per-paper configuration file. 
The rubric (Appendix~\ref{app:rubric}) and gate definitions are written in the prompts verbatim. Appendix~\ref{app:prompts} describes the prompts, and the pilot instructions are summarized in Appendix~\ref{app:pilot}.

\paragraph{Nature of human oversight and verification.} Humans selected the paper pool and the sampling rule, reviewed each configuration file before its run began (the config fixes the reproduction targets and the conclusion statement against which survival is defined, the pipeline's most consequential human input, bounded by the requirement that every headline estimate carry an exact table location), and checked the final reports. No human altered agent estimates or verdicts, and automation risk is bounded by gates decided by an agent other than their author and by withholding all destructive-action permissions (packages the prompts forbid agents to modify, and a separate output directory per run). 

\paragraph{Limitations and failure modes encountered.} Model style biases documented in the pilot imply assessment outputs are not model-invariant; the rubric constrains but does not eliminate judgment variation in fork verdicts. Agent knowledge cutoffs postdate the assessed papers' publication, so contamination by memorized paper content is possible in principle. However, the task we require the agents to do (checking possible forks that did not exist) is not in the existing papers, so the likelihood of contamination is low. 
Operational failures are part of the record (interrupted grid runs resumed from incremental checkpoints without data loss, timed-out watcher processes superseded by on-disk verification, API error due to various reasons), and every formalized assessment needed zero or one fix round at the code review gate. 
In the pilot, 153 agents were launched and 150 produced analyzable results. Two stopped at the budget cap and one produced no output. 

\paragraph{Experiences with hallucination and fabricated content.} The implementation reviews serve as a systematic log of fabricated or erroneous agent output, because every assessment's code and claims were re-derived by an independent agent. Across the ten implementation reviews, five blocker-level and forty-one major findings were recorded in their severity tallies, a count the manuscript's statistics script reads back from the review files. Among them, one agent mislabeled an invalid permutation procedure as a randomization test, and another built a placebo test that the true effect contaminated. Others were silent scope substitutions; the rest were ordinary engineering bugs.

\section{Pilot Study Details}
\label{app:pilot}

\subsection{Protocol and cost}

Each pilot agent is an instance of Claude Code running fully autonomously inside a Singularity container (40~GB RAM, 6 CPUs, \$20 budget cap per stage) with read-only access to the TAQ Parquet store. Instructions ask the agent, for each hypothesis, to propose and motivate a measure, estimate the average per-year change, test against the null of no change, and report the estimate with its standard error. The three-stage protocol, peer-evaluation design, top-paper selection, effect-size conversion pipeline, and full agent instructions follow the design summarized in Section~\ref{sec:pilot}. 

The total cost was \$1,558 across 450 agent-stage runs (median \$3.17 per agent for Stage~1). Median wall-clock time was 53 minutes per Stage-1 agent, and the pilot took about 25 hours end-to-end at 12-way parallelism.

\subsection{Additional pilot results}

Three pilot results are referenced in the text. (i) \emph{Empirical styles}: the Anderson--Darling test rejects equality of Sonnet and Opus effect-size distributions for all six hypotheses ($p \le 0.001$, the test's tabulated floor). 
The largest gaps are measure choice for market efficiency (autocorrelation: 87\% of Sonnet agents vs.\ 0\% of Opus) and functional form (log specification on H1, H2, and H5: 4--31\% of Sonnet vs.\ 64--88\% of Opus). (ii) \emph{Peer-review movement without convergence}: between Stages 1 and 2, 42\% of agents change measures and 44\% move their estimate by more than 0.5\%/yr, yet the stage IQR changes by less than 11\% for every hypothesis; for the two hypotheses whose dispersion is generated by a measure-family split, movements toward and away from the cross-agent median are balanced (H1 39\%/43\%, H4 41\%/40\%), and pooled across the six, 28\% of agent-hypotheses move toward the median and 36\% away. (iii) \emph{Cross-vendor replication}: 36 OpenAI Codex CLI agents (18 on \texttt{gpt-5.2-codex}, 18 on \texttt{gpt-5.3-codex}, through an Azure OpenAI enterprise endpoint) ran the identical Stage-1 protocol in the same containers, and 35 completed. The same conversion pipeline processed their outputs (one H6 estimate was unconvertible: the agent's signed-volume merge produced an all-zero regressor, an agent bug recorded in the conversion log). Stage-1 IQRs are uniformly smaller than the Claude population's (H2: 0.08 vs.\ 0.43; H4: 1.69 vs.\ 10.69\%/yr), medians agree where the construct is pinned (H2: $-6.17$ vs.\ $-6.21$; H5: $+2.93$ vs.\ $+3.08$\%/yr), and the fork taxonomy coincides (measure family and absolute-versus-signed on H1, trade weighting on H3, measure type on H6) while the fork mix differs (25 autocorrelation vs.\ 9 variance-ratio agents on H1; 34 of 35 agents on share volume for H4).

\subsection{Pilot agent instructions (abridged)}

Every pilot agent receives the same two fixed markdown files. 
1) an environment file (data schemas, resource limits, deliverable formats) and 2) an instruction sheet with the research context and the six hypotheses. We show the prompts below. Elisions are marked \texttt{[...]}, file-system paths are generalized, the project codename is replaced. The environment file also carries the full column schemas of the trades and NBBO Parquet files.

\begin{promptshade}
\promptname{Stage 1: Research analyst}
\begin{verbatim}
You are a research analyst in the [project] project. Your task is to
independently test six hypotheses about market microstructure trends in
SPY using NYSE TAQ data. [...]
ASSIGNMENT: You are expected to write a research report (2,000-4,000
words). For each of the six hypotheses listed below, you should:
1. Propose a statistical measure, briefly motivate it, and present the
   formula to calculate it.
2. For this measure, estimate the average per-year change in percentage
   points, based on the full sample (or at least the longest possible
   period [...]). Test it against the null of no change.
3. Report this estimate along with its standard error and the direction
   of the effect. Report at least three digits after the decimal point.
4. Briefly discuss your result.
All results are in percentage points per year (e.g., a 1.234% annual
decline is reported as -1.234, not -0.01234).
HYPOTHESES (each stated as a null of no change over 2015-2024):
H1 Market Efficiency (random-walk benchmark); H2 Quoted Bid-Ask Spread;
H3 Realized Bid-Ask Spread; H4 Trading Volume; H5 Intraday Volatility;
H6 Price Impact of Trades. [...]
DELIVERABLES: results.csv with exactly 6 data rows (hypothesis,
measure_name, effect_size, standard_error, t_statistic, direction,
n_periods, frequency, model, software); report.md; all code.
HARD CONSTRAINTS: the TAQ Parquet store is read-only (66 GB, over 7
billion rows); do not load all files into memory at once.
\end{verbatim}
\end{promptshade}

\begin{promptshade}
\promptname{Peer evaluator (two independent instances per Stage-1 report)}
\begin{verbatim}
For each report assigned to you (located in review/), you should:
1. Rate the overall quality of the paper on a 0-10 scale.
2. Rate each hypothesis (H1 through H6) on a 0-10 scale based on the
   quality of the analysis.
3. Provide written feedback [...]. For each hypothesis and for the paper
   overall, focus on what you see as central weaknesses in the chosen
   econometric strategy. Provide clear and concise reasons why you see
   your proposed revisions as material to the quality of the analysis.
DELIVERABLES: evaluation_N.md and ratings_N.csv per assigned report.
\end{verbatim}
\end{promptshade}

\begin{promptshade}
\promptname{Stage 2: Revision under peer feedback}
\begin{verbatim}
You previously submitted a research report testing six hypotheses about
market microstructure trends in SPY (Stage 1). Your report has been
evaluated by two anonymous peer evaluators who have rated your work and
provided written feedback. Please submit a revised version of your
report. [...]
WHAT YOU HAVE: your Stage-1 report, results, and code;
feedback/evaluation_1.md + ratings_1.csv; evaluation_2.md +
ratings_2.csv.
DELIVERABLES: results_v2.csv, report_v2.md, code_v2/.
\end{verbatim}
\end{promptshade}

\begin{promptshade}
\promptname{Stage 3: Revision after exemplar exposure}
\begin{verbatim}
You have previously submitted a research report (Stage 1) and a revised
version after receiving peer feedback (Stage 2). You have been given the
five short papers that received the highest average peer evaluation
ratings. Please read them and thereafter submit another revised version
of your report. [...]
WHAT YOU HAVE: your Stage-2 report, results, and code;
top_papers/paper_1.md + results_1.csv ... paper_5.md + results_5.csv
(anonymized).
DELIVERABLES: results_v3.csv, report_v3.md, code_v3/.
\end{verbatim}
\end{promptshade}

\subsection{Complete pilot specification curves}

\begin{figure}[H]
\centering
\includegraphics[width=\textwidth]{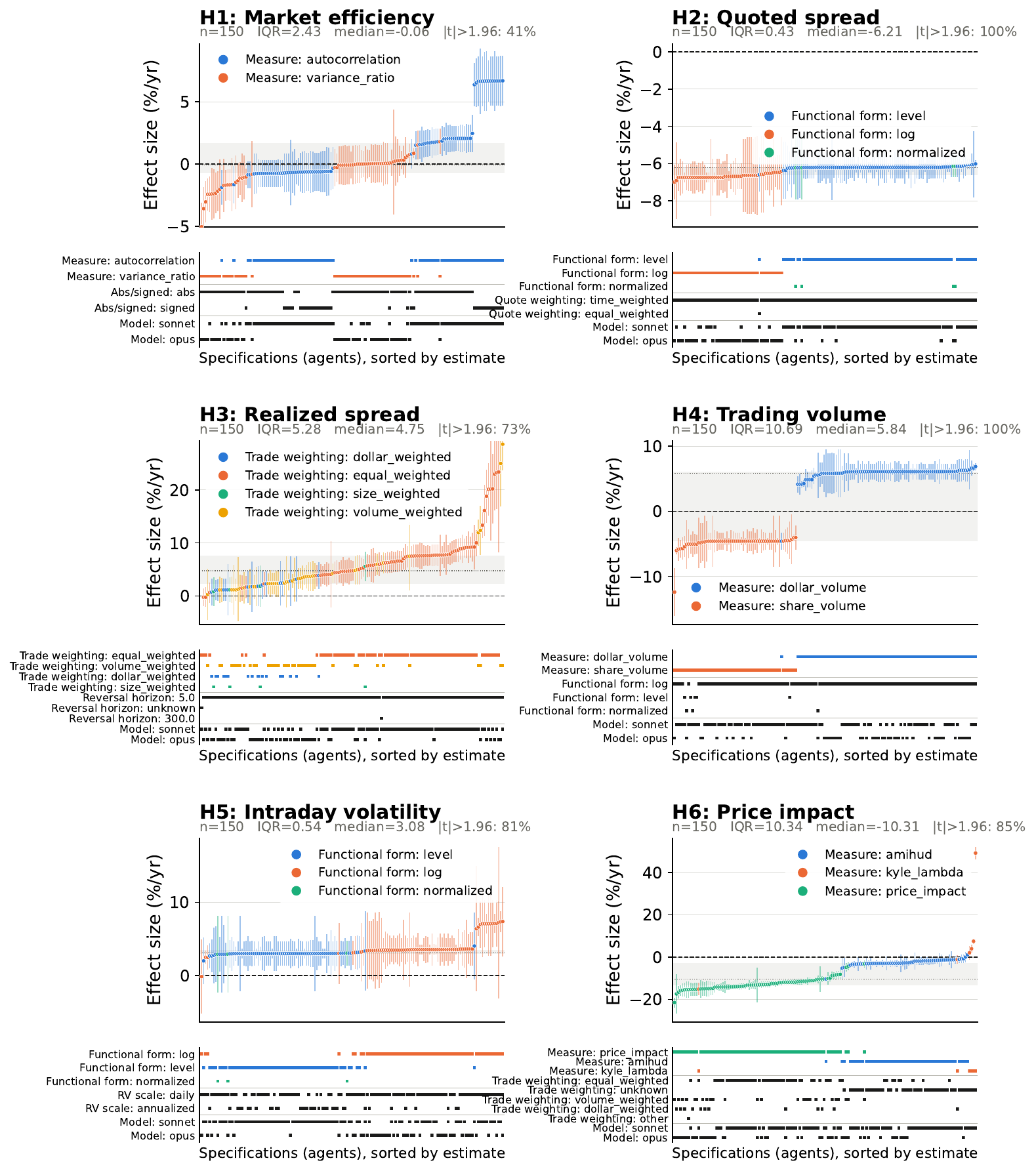}
\caption{Complete Stage-1 specification curves, all six hypotheses (150 agents each): dispersion concentrates on a handful of discrete forks, visible as contiguous color blocks in the fork matrices.}
\label{fig:pilot_full_s1}
\end{figure}

\begin{figure}[H]
\centering
\includegraphics[width=\textwidth]{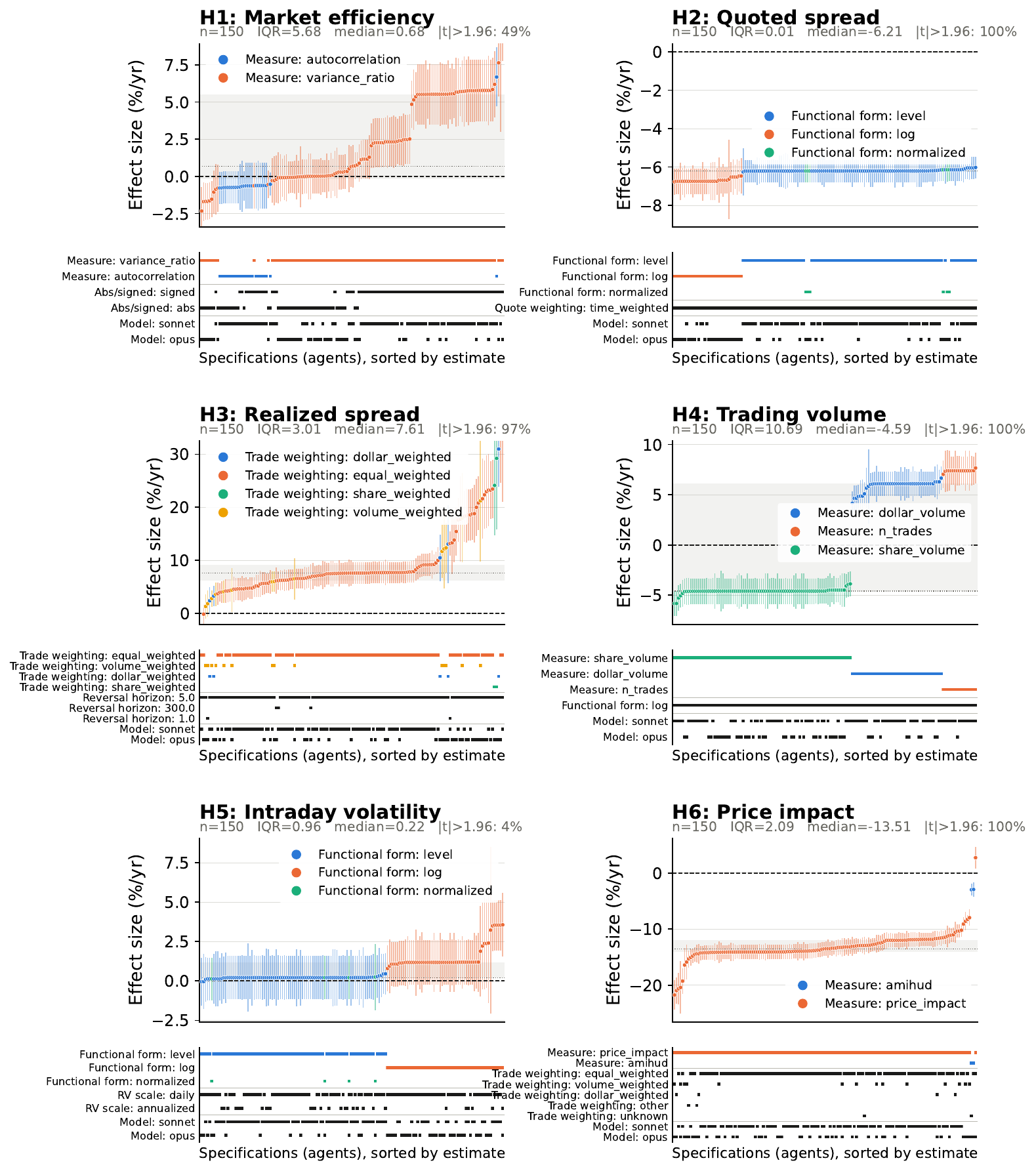}
\caption{Complete Stage-3 specification curves (after exemplar exposure): within-family dispersion collapses by 80--99\% through imitation of the exemplars' choices, not through analytical convergence.}
\label{fig:pilot_full_s3}
\end{figure}

\begin{figure}[H]
\centering
\includegraphics[width=\textwidth]{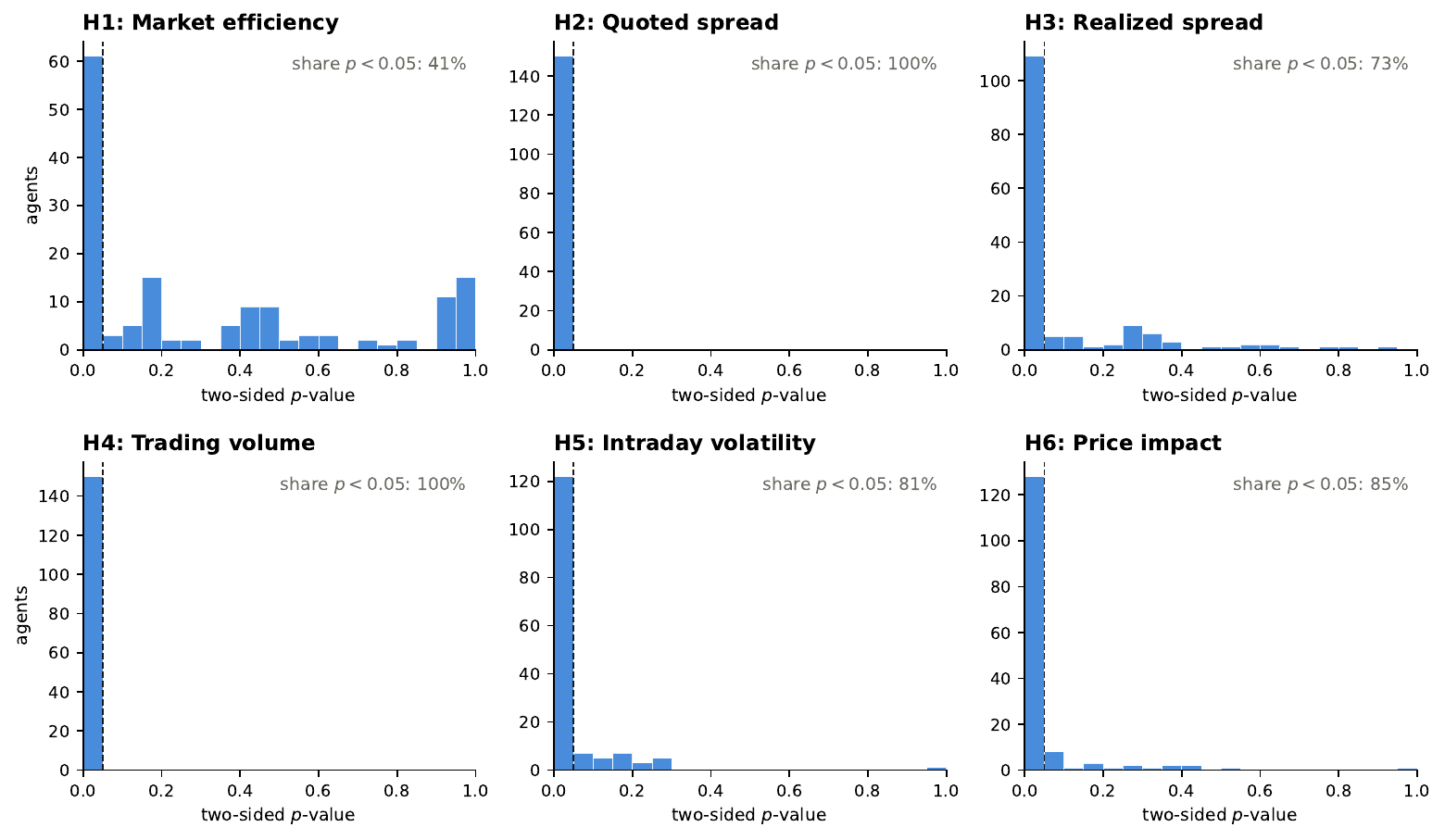}
\caption{Distribution of two-sided $p$-values implied by each agent's converted Stage-1 $t$ statistic (normal approximation), 150 agents per hypothesis; dashed line at $p = 0.05$. High significance shares coexist with cross-agent disagreement: for trading volume every estimate is significant while the estimates split across opposite signs by measure family.}
\label{fig:pilot_pvals}
\end{figure}

Figure~\ref{fig:pilot_full_s1} draws the complete Stage-1 specification curve for every hypothesis. All 150 agents sorted by converted estimate with 95\% intervals, and a fork-choice matrix beneath each panel (measure family, functional form, the hypothesis-specific forks, and the model family). Figure~\ref{fig:pilot_spec} is the H1 panel of this figure restricted to its four highest-invalidation-rate forks. Figure~\ref{fig:pilot_full_s3} repeats the drawing for Stage~3. After seeing the exemplar, the measure-family blocks that generate the Stage-1 dispersion collapse through imitation (Section~\ref{sec:pilot}). Figure~\ref{fig:pilot_pvals} adds the distribution of two-sided $p$-values implied by each agent's converted Stage-1 $t$ statistic (normal approximation): significance is abundant even where agents disagree, most clearly for trading volume, where every estimate clears $p<0.05$ while the population splits across opposite signs, so the pilot's dispersion is analytical disagreement instead of sampling noise. 

\section{The Fork-Validity Rubric}
\label{app:rubric}

The fork auditor scores every brainstormed option on the three conditions of \citet{simonsohn2020specification}. 

\emph{S1, construct consistency}: the option estimates the same theoretical quantity the paper's headline claim is about (an option that changes the estimand, e.g., a per-unit versus an aggregate version of the same outcome, is retained but labeled as a distinct estimand, but will not be calcuated in survival).

\emph{S2, statistical validity}: the option's estimator and inference are appropriate to the design (options that cluster below the assignment level, condition on post-treatment variables, or compare against already-treated units are rated invalid as \emph{primary} specifications, though they may be retained as labeled diagnostics). 

\emph{S3, non-redundancy}: options that are algebraic relabelings of other options are merged. Each verdict carries a written reason. The grid is the cross-product of surviving options, capped at 100--250 cells by auditing marginal informativeness. The auditor also designates, before estimation, the binding inference method per Section~\ref{sec:framework}, and flags mandatory companion diagnostics (placebo tests, balance checks) that must disclose with every headline cell.

\section{Gate Definitions}
\label{app:gates}

\textbf{G0 (reproduction).} Published headline coefficients reproduced within 15\% (in practice they are reproduced almost exactly).  \textbf{G2 (fork coverage).} Every brainstormed option carries an verdict, with a stated reason. The inference method for $p$-value is also specified., 
\textbf{G4 (implementation).} Zero unresolved blocker findings after the fix-and-re-audit loop; every fix is itself re-audited. 
\textbf{G6 (report integrity).} Every numeric claim in the report is generated by an audited scripts. Self-reported numbers will fail the gate. 

In the audits reported here, G0 passed for every completed audit (with the targets of Section~\ref{sec:framework} and Table~\ref{tab:repro} for the three packages that did not include result logs). The implementation reviews recorded five blocker-level and forty-one major findings in their severity tallies across the ten assessments, each resolved or explicitly adjudicated before results were read, with every audit needing zero or one fix round at G4.

\subsection{Measured reproduction precision}

Package estimation code is mostly Stata. In the ten original runs no Stata is executed (the main version only tested open-source softwares), so every G0 reproduction in Table~\ref{tab:repro} is an independent re-implementation in Python or R. As a robustness check, the second vendor's full-framework run (Section~\ref{app:xvendor}) did use Stata. G0's 15\% translation tolerance was not needed, the measured deviations sit at the level of published rounding. Table~\ref{tab:repro} summarizes the per-paper records.

\begin{table}[H]
\centering
\caption{Measured reproduction precision by paper: the largest observed deviation on any G0 target. Target types: published table values; the authors' own result log; or, for packages shipping neither, the package-output analog (the numbers the authors' code produces on the shipped data, Section~\ref{sec:framework}).}
\label{tab:repro}
\footnotesize
\begin{tabular}{cllp{6.2cm}}
\toprule
Paper & Grid & G0 targets & Measured deviation \\
\midrule
A & Python & published & estimates and SEs exact to reported digits \\
B & R & published & coefficients, SEs, and $N$ exact (0.0\%) \\
C & Python & published & headline coefficients match within published rounding; a small $N$ gap is explained by missing-value handling \\
D & R & published (authors' log) & linear columns to 7 significant digits; nonlinear columns within 0.04\% (estimator translation); $N$ exact \\
E & Python & package analog & coefficients within $5\times10^{-7}$ (exact at reported digits) \\
F & R & package analog & coefficients, clustered SEs, $N$ exact to 4 decimals \\
G & Python & published & coefficients and $N$ exact; one SE $+0.04\%$ (fixed-effects convention) \\
H & Python & authors' log & exact at printed decimals; $N$ exact \\
I & Python & package analog & deviations within published rounding; $N$ exact \\
J & Python & published & within published rounding \\
\bottomrule
\end{tabular}
\end{table}

\section{Decomposing Non-Robustness}
\label{app:decomp}

A conclusion can fail because its \emph{sign} is not robust (the published direction is a minority outcome across defensible specifications), because its \emph{magnitude} is not robust (the direction holds but the defensible effect size is materially smaller), because its \emph{significance} is not robust (sign and magnitude hold, but $p<0.05$ fails under the designated inference), because it fails \emph{identification diagnostics} (sign, magnitude, and significance all hold, yet a placebo or balance test attributes the estimate to something other than the treatment), or because it is \emph{estimand-dependent} (different defensible operationalizations of the claim answer different questions, the distinction \citealp{shmueli2010explain} draws between the quantity a model explains and the quantity it predicts). The audits distinguish all five, and the last two do matter. Paper J passes the first three axes and fails only the fourth, and Paper D fails only the fifth, so a decomposition restricted to sign, magnitude, and significance would misclassify both.

Sign and significance are the manuscript's main metrics. Table~\ref{tab:decomp} in addition, analyzes the magnitude dimension,  For each paper, the median and interquartile range of the grid estimate as a ratio of the published estimate, are comparable to the published specification. 

\begin{table}[H]
\centering
\caption{Decomposition of robustness failures across the ten assessments: the sign, magnitude, and significance axes separate cleanly. Magnitude is the median [IQR] of cell$/$published estimate over the $n$ comparable cells (1 = published magnitude, 0 = null, negative = sign flip; filters documented in the script).}
\label{tab:decomp}
\footnotesize
\resizebox{\textwidth}{!}{
\begin{tabular}{lccclc}
\toprule
Paper & Sign (\%) & Magnitude: median ratio [IQR] ($n$ cells) & Survive (\%) & Identification diagnostics & Binding axis \\
\midrule
A & 100 & 0.50 [0.42, 0.85] (96)$^{a}$ & 30 & placebos pass (three falsification checks); curve permutation $p = 0.035$ & magnitude (revision only) \\
B & 72 & 0.32 [0.05, 0.48] (81)$^{b}$ & 9 & --- & significance (headline leg) \\
C & 95 & 1.27 [0.82, 1.55] (88) & 3 & --- & significance (assignment-level inference) \\
D & 99 / 40 & 0.54 [0.38, 0.66] (48)$^{g}$ & 78 / 7 & secondary-outcome selection bounds never exclude zero & outcome family (secondary outcome) \\
E & 22 & $-0.27$ [$-0.61$, $-0.02$] (36)$^{c}$ & 0 & effect traced to already-treated comparisons & sign \\
F & 100 / 68 & n.q.$^{d}$ & 67 / 20 & secondary-design functional-form variant unidentified & (primary design robust); secondary-design significance \\
G & 95 & 0.60 [0.26, 1.01] (110)$^{e}$ & 17 & data-construction artifact corrected & significance (marginal) \\
H & 77 & 0.88 [0.60, 1.14] (18) & 0 & published inference's implicit placebo centered away from zero & significance (under valid inference) \\
I & 64 & 0.07 [$-0.03$, 0.27] (48)$^{f}$ & 10 & comparison group shifts at treatment onset & magnitude and significance \\
J & 93 & 0.54 [0.19, 0.95] (64) & 47 & pre-outcome placebo fails in 90 of 91 survivors & identification diagnostics \\
\bottomrule
\end{tabular}
}
\begin{minipage}{\textwidth}
\vspace{4pt}
\footnotesize Sign and survival repeat Table~\ref{tab:overview}. $^{a}$Published-specification cells, de-duplicated. $^{b}$Primary-model cells, proportional-effect scale. $^{c}$The sign failure seen from the magnitude axis. $^{d}$Scales differ across F's families. $^{e}$Vs.\ the median of the published variants. $^{f}$Full comparable universe; the trend-corrected arms cohere at $\sim$1/4--1/3 (Section~\ref{sec:audits}). $^{g}$Primary-outcome cells; sign and survival shown for the primary/secondary outcome.
\end{minipage}
\end{table}

The results show a more diverse result. First, the axes separate cleanly in practice: H is a pure significance failure (magnitude ratio 0.88, centered near the published value), E is a sign failure (the median comparable cell has the opposite sign), and I is primarily a magnitude failure (median ratio 0.07 over the full comparable universe). Second, the magnitude axis adds information the survival in the main text does not have: B's finding loses roughly two-thirds of its effect size in the median comparable cell, while H's median effect size is the closest to the published value in the sample even though its survival is zero. 

\section{Specification Curves for the Assessed Papers}
\label{app:ecspec}

\begin{figure}[!htbp]
\centering
\includegraphics[width=\textwidth]{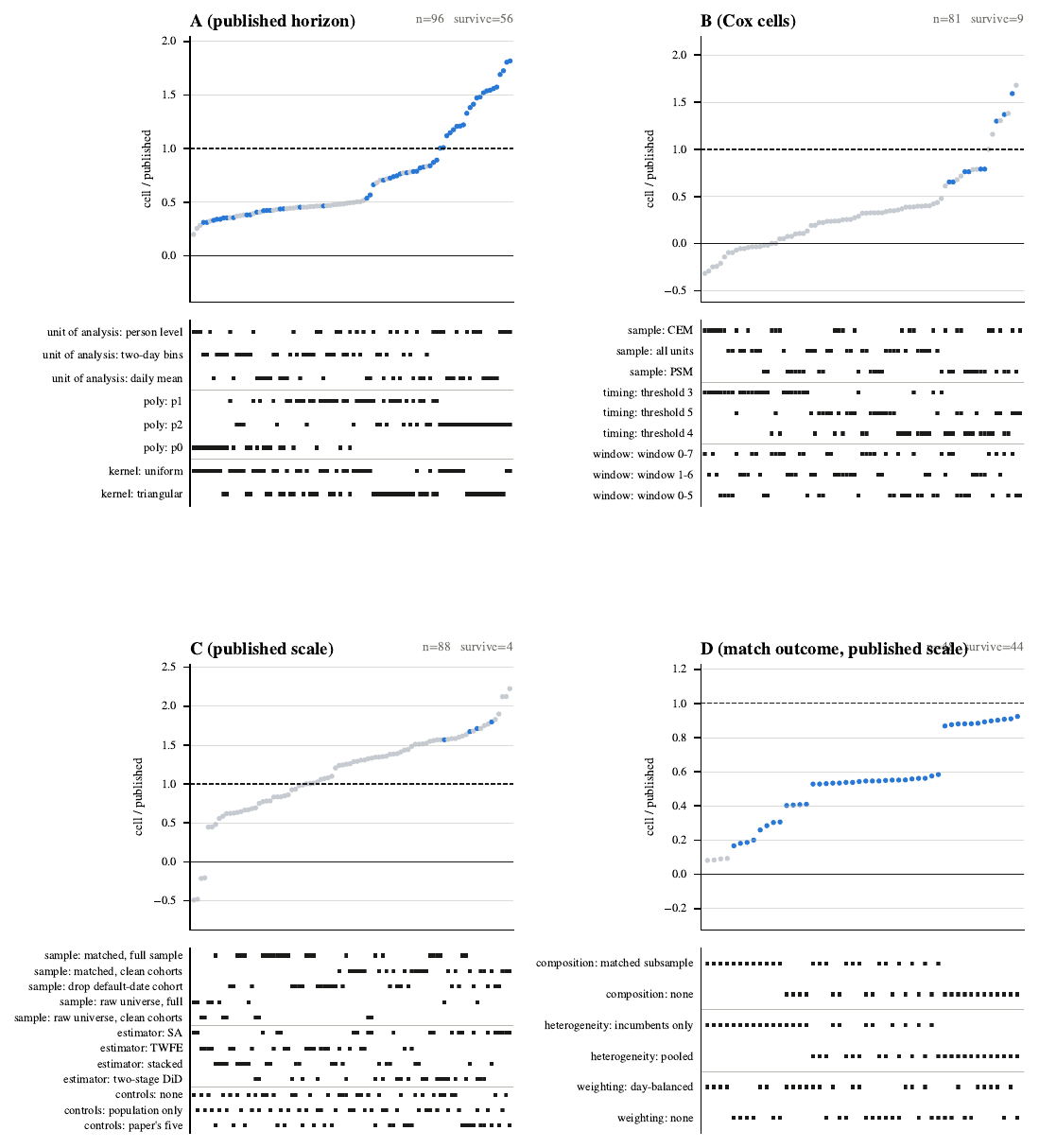}
\caption{Specification curves for the assessed papers with a unit-comparable published scale (I of III: Papers A, B, C, and D, match outcome), as cell$/$published ratios: dashed line = published value, solid line = zero, dark = survives the designated inference within the panel's subset. Beneath each curve, a fork matrix marks every specification's option on the panel's three highest-invalidation-rate dimensions (labels generalized to preserve anonymity).}
\label{fig:ec_spec}
\end{figure}

\begin{figure}[!htbp]
\centering
\includegraphics[width=\textwidth]{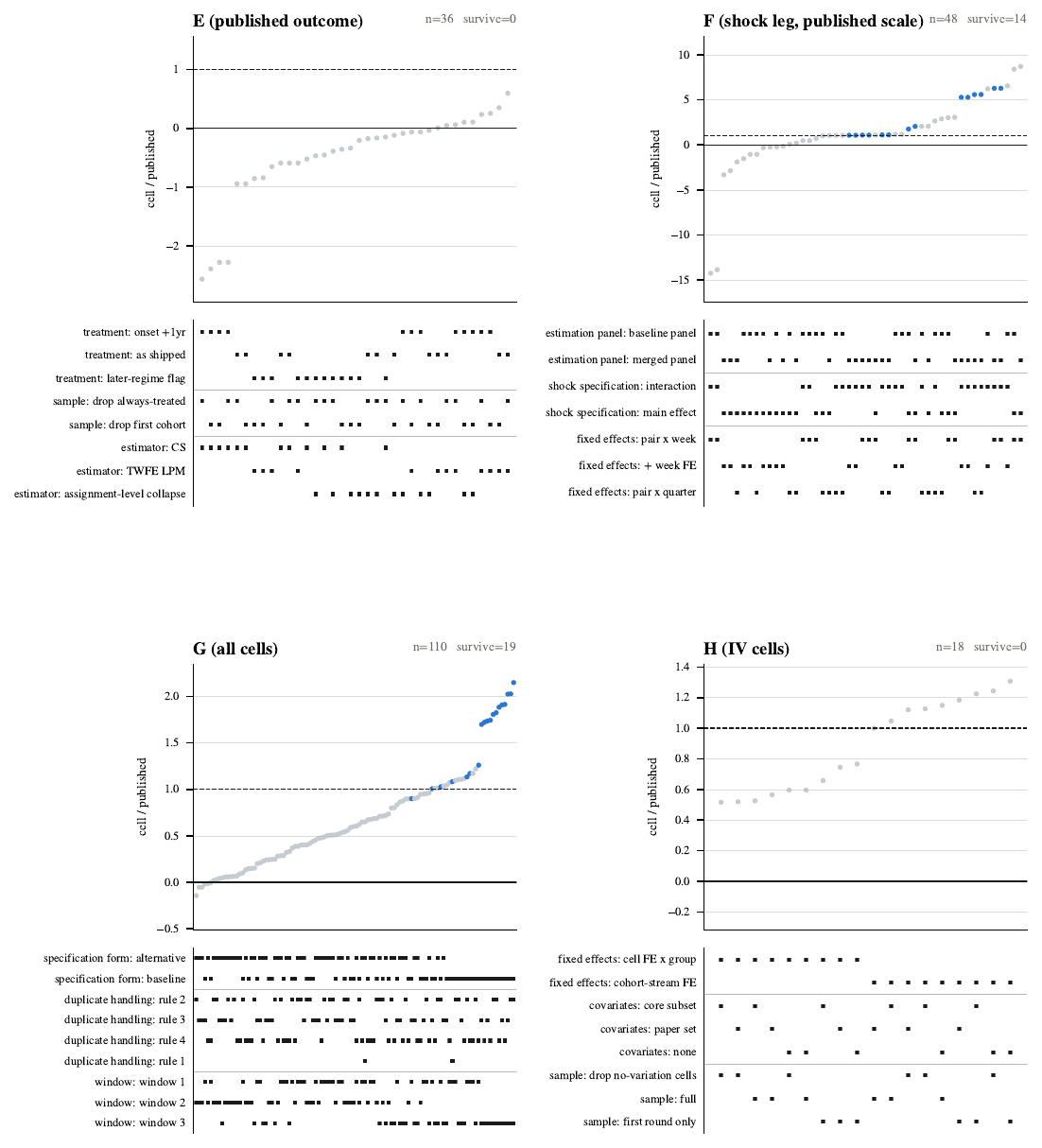}
\caption{Specification curves, continued (II of III: Papers E, F (shock leg), G, and H).}
\label{fig:ec_spec2}
\end{figure}

\begin{figure}[!htbp]
\centering
\includegraphics[width=.72\textwidth]{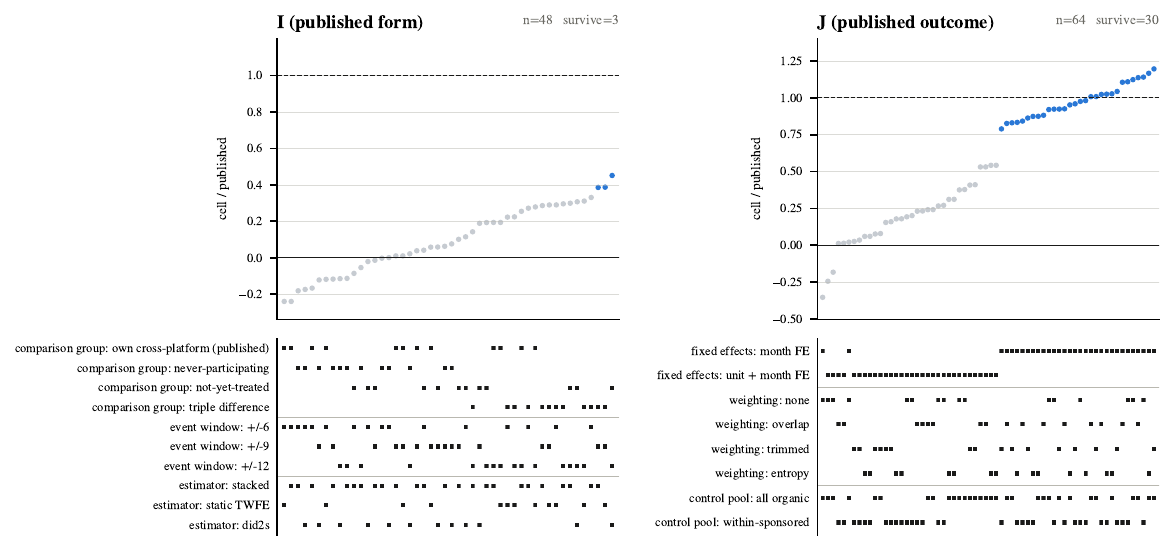}
\caption{Specification curves, continued (III of III: Papers I and J).}
\label{fig:ec_spec3}
\end{figure}

Figures~\ref{fig:ec_spec}--\ref{fig:ec_spec3} draw the specification curve for each assessed paper with a unit-comparable published scale, plotted as the ratio of each cell's estimate to the published estimate so that no absolute value appears. All ten papers have such a panel.  Paper A is normalized to its published mean effect on the published-specification cells, and Paper D to its published primary-outcome coefficient (its secondary-outcome family has no stable published scale and is scored in Table~\ref{tab:overview} only). For Paper F only the secondary design is drawn: the primary design's published coefficients are not unit-comparable to the audited grid, whose data-cleaning choices change the estimation sample's scale (its survival is scored on its headline block in Table~\ref{tab:overview}). The panels use the magnitude-comparable subsets of Section~\ref{app:decomp} (G uses all cells; F uses the continuous-scale secondary-design cells), so survivor counts can differ from Table~\ref{tab:overview}, which scores each paper's full universe, and J's survivors ignore the placebo diagnostic that only one of them passes. Beneath each curve a fork-choice matrix marks, for every specification, its option on the panel's three highest-invalidation-rate dimensions, so we can observe the decisive forks. 

\begin{figure}[!htbp]
\centering
\includegraphics[width=.66\textwidth]{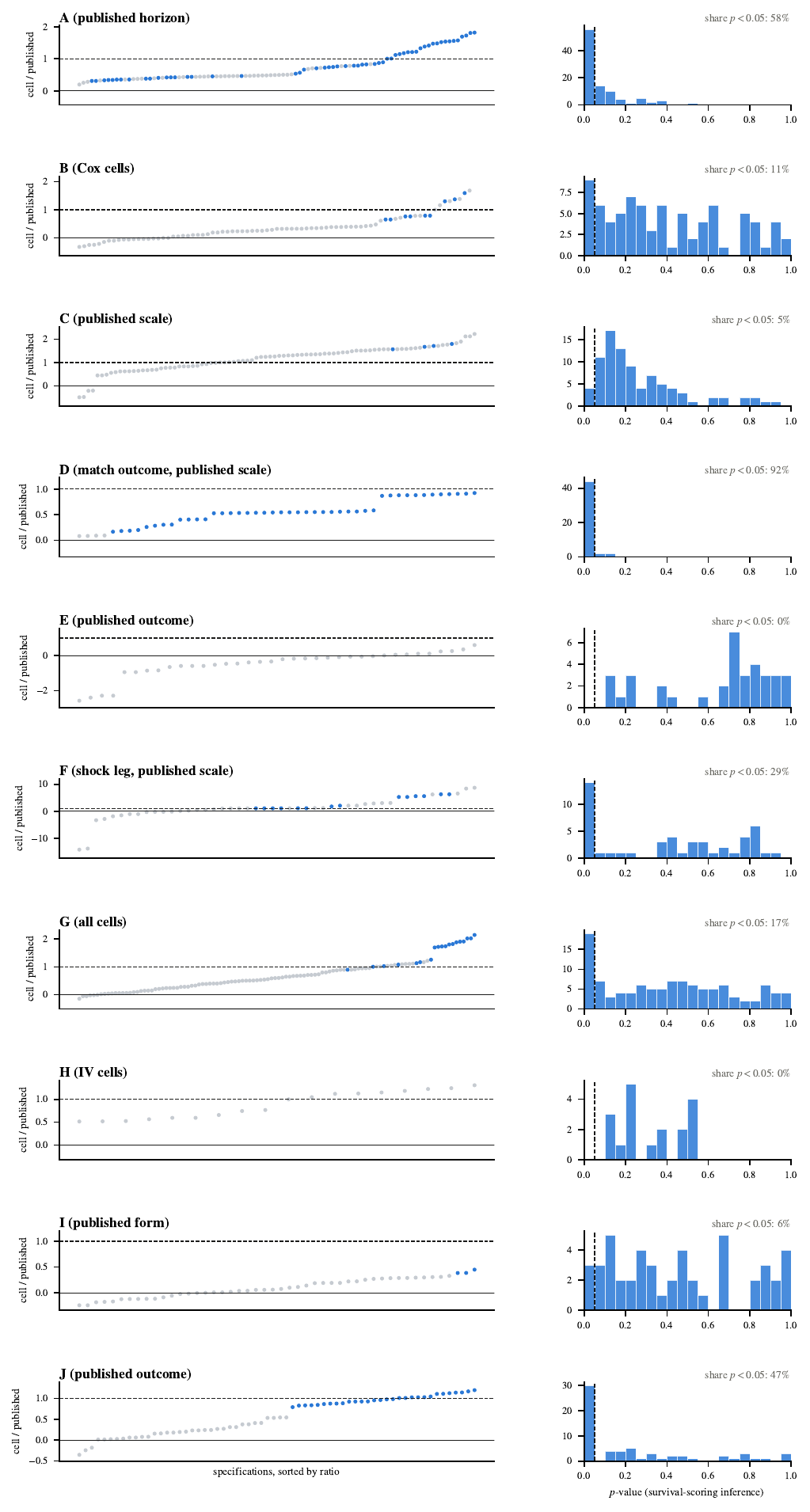}
\caption{Effect sizes and $p$-values side by side for the ten papers. Left: the specification curve as cell$/$published ratios (dark = survives). Right: the distribution of the $p$-values used for survival scoring in that panel, under its designated inference; dashed line at $p = 0.05$, with the share below it at top right. Generated by the committed script \texttt{make\_ec\_spec\_curves.py}.}
\label{fig:ec_spec_pval}
\end{figure}

Figure~\ref{fig:ec_spec_pval} places each panel's effect sizes and $p$-values side by side. The ratio curve next to the distribution of the $p$-values that score survival in that panel, under its designated inference method. The published papers' valid universes concentrate $p$ mass above 0.05.

\section{Implementation Details and Role Prompts}
\label{app:prompts}

\subsection{Data scale and computational environment}

Headline samples range from hundreds to about a million observations, and processed data from about 10~MB to several GB per package; per-paper values are withheld to preserve anonymity. Package code is mostly Stata. The ten original grids re-implement estimation in Python or R; the second vendor's grids (Section~\ref{app:xvendor}) used Stata for two papers.

\paragraph{Computational environment.} All Claude agents run as Claude Code CLI sessions on a shared, Slurm-managed university HPC cluster, and the second vendor's and the third model's agents run through the Codex CLI on the same cluster; every workload is CPU-only. Pilot agents execute inside Singularity containers built from a pinned Docker image (Python 3.12 plus the agent CLI), each with 40~GB RAM and 6 CPUs, up to 12 in parallel, with the TAQ data mounted read-only and a private workspace per agent, the isolation that guarantees cross-agent independence (Section~\ref{sec:pilot}). 

Framework assessments run in-process with a 16~GB memory cap per grid execution. 
Estimation uses a pinned stack: Python with \texttt{pyfixest}~0.60 (including its \texttt{did2s} implementation), \texttt{rdrobust}, \texttt{wildboottest}, and \texttt{pyreadstat}, R~4.2.2 inside the same container image with \texttt{fixest}, \texttt{did}, \texttt{DRDID}, and \texttt{clubSandwich}, used whenever a specification needs fixed-effects GLMs on large panels. Each run works on its own copy of the replication package, which the prompts forbid agents to modify, and every agent session is captured as a structured log with its model identifier, prompt, tool calls, and cost, which is the assessment trail behind the GenAI Use Appendix.

This appendix documents how the pipeline of Section~\ref{sec:framework} is actually run and reproduces the role prompts in abridged form. The excerpts below are verbatim from the versioned templates except for four normalizations: file-system paths are generalized, the pipeline's internal codename is replaced by ``the audit pipeline,'' the identifier of the audited journal's package archive is redacted for anonymity, and typographic dashes are normalized; elisions are marked \texttt{[...]}.

\subsection{Orchestration contract}

One run is one paper with a replication package. An orchestrator session renders each role's fixed prompt template, filling \texttt{\{\{PLACEHOLDER\}\}} fields from the paper's \texttt{config.yaml}, launches the role as an independent agent session in the fixed order C, 0, 1, 2, 3, 4, 5, 6, and checks the gate artifacts between steps; a failed gate stops or loops the run. Gate checking belongs to the orchestrator, not to the gated agent: the reproducer's printed verdict line (visible in its prompt below) is a convenience summary, and the G0 decision is the orchestrator's comparison of the reproduction table against the configured targets; G2's coverage check is programmatic; G4's decision is the independent auditor's re-audit of every fix and rebuttal; G6's is the requirement that the report be emitted by a committed generator script. 
Section~\ref{sec:framework} states each gate's verification status and its residual weaknesses. 

\subsection{Per-paper configuration}

The configuration file records the paper's identity, the one-sentence causal research question as the paper states it, the published conclusion (direction, plus every headline estimate the reproduction must hit, each with its exact table location), the design in two to four sentences (unit, sample, treatment, estimator, inference as implemented in the code), the authors' own fork choices in dimension:option form (starred in the report), all data paths the agents may touch, with notes on known quirks; and the compute block (interpreter paths and the 16~GB memory cap). A config-drafting agent proposes this file from the package alone, and a human reviews it before the run starts. 

\subsection{Role prompts (abridged)}

\begin{promptshade}
\promptname{Step C: Config drafter}
\begin{verbatim}
You are the CONFIG DRAFTER for the audit pipeline. [...] PACKAGE: {{PKG_DIR}}
(a [redacted journal volume-year] replication package; paper identity may only
be inferable from readme/PDFs/code comments/pre-registrations inside).
TASK: Identify the paper (title, authors, venue+year) and draft a complete,
accurate config.yaml. [...] If the paper's published coefficients are visible
(log files, readme tables, PDF), record them as headline_estimates with exact
locations; if not, record the estimates that the package's own code produces
for the MAIN table [...] and mark where: "package output (Table X analog)".
[...] If the paper is NOT a reduced-form empirical paper (pure simulation/
theory/structural-only), say so and stop.
\end{verbatim}
\end{promptshade}

\begin{promptshade}
\promptname{Step 0: Reproducer}
\begin{verbatim}
You are the REPRODUCER in a fixed audit pipeline. [...] Do not modify any
package file. TARGETS - reproduce each of these published numbers:
{{HEADLINE_ESTIMATES}}
TASK: Decode the design from the package's code/readme (read the actual
estimation code, not just prose), then re-implement the headline estimation
yourself [...] and reproduce every target. Where your implementation differs
from the package's (e.g., library versions), quantify and explain the residual.
OUTPUT: [...] (1) the decoded design in <=10 lines (unit, sample filters,
outcome, estimator, inference - exact enough that a stranger could
re-implement); (2) a table: target | published | reproduced | deviation;
(3) verdict line "GATE0: PASS" if all coefficients within 15% AND design
quantities (bandwidths, N, cohort dates) exact, else "GATE0: FAIL - <why>".
\end{verbatim}
\end{promptshade}

\begin{promptshade}
\promptname{Step 1: Fork brainstormer}
\begin{verbatim}
You are the FORK BRAINSTORMER in a fixed audit pipeline. [...] You enumerate;
a separate auditor judges. Do NOT assess validity.
TASK: For each DESIGN CHOICE (dimension) a competent researcher faces when
answering this research question with these data, list ALL plausible options -
including ones you suspect are wrong. Cover at minimum: outcome
operationalization(s), sample definition, treatment/running-variable coding,
estimator family, functional form, controls, fixed effects, weighting,
inference, time window, unit of analysis; add paper-specific dimensions you
see. Every option must be implementable with the listed data - name exact
files/columns/formulas. Verify empirical claims in the data rather than
assuming (record what you verified in a notes_on_data block).
\end{verbatim}
\end{promptshade}

\begin{promptshade}
\promptname{Step 2: Fork auditor}
\begin{verbatim}
You are the FORK AUDITOR in a fixed audit pipeline - independent of the
brainstormer; be adversarial and cite standard econometrics [...].
READ: pipeline/rubric.md (fixed rubric: S1 construct-consistent, S2
statistically valid, S3 non-redundant) [...]. You may run small data checks
[...] to verify empirical claims that decide a verdict.
TASK 1: For EVERY option assign VALID / QUESTIONABLE / INVALID for HEADLINE
use, one-sentence reason, citation where standard. Judge the paper's own
choices with the same severity as any other.
TASK 2: Combination rules - jointly invalid or redundant sets.
TASK 3: recommended_grid: cross the VALID options of the 4-8 most
consequential dimensions (100-250 cells); every non-crossed dimension gets a
held-fixed entry {dimension, option_id, why}; specify the admissible INFERENCE
method for headline p-values (this binds step 3).
\end{verbatim}
\end{promptshade}

\begin{promptshade}
\promptname{Step 3: Implementer}
\begin{verbatim}
You are the IMPLEMENTER in a fixed audit pipeline. [...] Your code will be
audited line-by-line by an independent agent; write for that reader.
TASK: Write run_grid.(py|R) implementing EXACTLY the recommended grid - no
silent substitutions: if an option is infeasible, record a row with
err="infeasible: <why>" rather than approximating. [...] Inference: implement
the audit's designated method (wild/block bootstrap, robust bias-corrected, or
randomization inference) - never plain analytic SEs if the audit forbade them.
Engineering rules (each has caused a real failure): write results
incrementally (one CSV row per finished cell, append); wrap each cell in
try/except recording the error; column-select on load, peak RAM <= 16 GB; if
projected wall-clock > 2h, downscale deterministically and document the rule
in the script header; never name a dataframe column after a library method;
when weights exist they must reach EVERY estimator call; verify date/id
columns parse as intended for every extra data file. Then RUN it to completion
and sanity-print: n cells, n errors, median/IQR of the estimate, share
positive, share significant under the designated inference.
\end{verbatim}
\end{promptshade}

\begin{promptshade}
\promptname{Step 4: Implementation auditor}
\begin{verbatim}
You are the IMPLEMENTATION AUDITOR in a fixed audit pipeline. You did not
write the code. Do not edit any file.
TASK: Verify the code implements the recommended grid faithfully. Hunt
specifically for: silent fork substitution (an option label whose code does
something else - the worst failure class); options dropped or approximated
without an err row; weights/covariates/filters computed but never passed to an
estimator; formula-parsing hazards; column-name collisions with library
methods; wrong date/id parsing in secondary data files (check dtypes
empirically); windows/cutoffs off by one; inference not matching the audit's
designation; [...] recompute 3 random cells independently and compare to
grid_results.csv.
OUTPUT: findings ranked BLOCKER (wrong estimates) / MAJOR (affects some cells
or interpretation) / MINOR, each with file/line, what is wrong, suggested fix;
end with an explicitly-cleared list (things you checked and found correct).
\end{verbatim}
\end{promptshade}

\begin{promptshade}
\promptname{Step 3F: Fix round (the Step-3 agent, resumed)}
\begin{verbatim}
You are the IMPLEMENTER, resumed for the FIX round. Read
implementation_audit.md. For every BLOCKER and MAJOR: fix the code, or rebut
in writing with evidence. Re-run whatever the fixes invalidate (full grid if
estimates change, affected cells otherwise). Log each finding's disposition
(fixed+rerun / rebutted+reason) in fix_log.md.
\end{verbatim}
\end{promptshade}

\begin{promptshade}
\promptname{Step 5: Author-commentary researcher}
\begin{verbatim}
You are the AUTHOR-COMMENTARY RESEARCHER in a fixed audit pipeline. [...]
identify the 3-6 forks that most threaten the published conclusion (lowest
survival / sign flips / INVALID rulings on the paper's own choices).
TASK: Find what the AUTHORS THEMSELVES say about each of those forks, in the
published paper or the closest obtainable version [...]. For each fork:
verbatim quote(s) with location, or "silent" - and then VERIFY testable claims
(if the paper claims a robustness "does not affect our results", re-run that
exact check on the package data and report the actual numbers). Classify each:
anticipated-and-defended / partially-anticipated / disclosed-but-downplayed /
silent / claim-contradicted-by-data.
\end{verbatim}
\end{promptshade}

\begin{promptshade}
\promptname{Step 6: Reporter}
\begin{verbatim}
You are the REPORTER in a fixed audit pipeline. [...] Every number in your
report must be COMPUTED by your script from the CSVs - never typed from
memory.
TASK: [...] Survival analysis: survive <=> sign(beta)==published direction AND
p<0.05 under the designated inference. Per fork option: n, survival rate, flip
rate; kill power per dimension (max-min survival). [...] a specification curve
with a fork-choice matrix beneath it, rows ordered by kill power, the paper's
own choices starred.
\end{verbatim}
\end{promptshade}

\subsection{Resource discipline}

Every agent session is capped at 16~GB of peak memory, with column-selected and chunked data loading mandatory for wide files. 
Long grid executions run in the background and write one CSV row per completed cell, so an interrupted run resumes without loss. 

\subsection{Cross-model replications}
\label{app:xvendor}

Here we report how AMA changes with different AI models. We discuss the cross-vendor replication of the pilot's Stage~1 and a cross-vendor replication of the complete assessment framework on all ten papers, and a third frontier model re-running the fork validation alone on four of them. 
\paragraph{The pilot's cross-vendor check.}
Thirty-six OpenAI Codex CLI agents (18 on \texttt{gpt-5.2-codex} and 18 on \texttt{gpt-5.3-codex}) ran the identical Stage-1 protocol in the same containers against the same data, and 35 completed; the same conversion pipeline processed their outputs. The fork taxonomy coincides with the Claude population's, and the medians agree wherever the construct is pinned, while the mix of options chosen within each fork differs enough to shrink every Stage-1 interquartile range. Section~\ref{app:pilot}, item~(iii), reports the numbers. The dispersion the pilot documents is therefore not an artifact of one vendor's models, but its magnitude is model-dependent.

\paragraph{The second vendor through the complete framework.}
The second check ran a second vendor's model (OpenAI \texttt{gpt-5.6-sol} through the Codex CLI at medium reasoning effort) through the complete framework on all ten papers. 
Every reproduction anchor passed and every paper cleared the code review gate. 
At the option level the two enumerations agree on 221 of 334 matched options (66\%), with 17 flat valid/invalid reversals. The comparison anchors on the options our fork validation placed in the dimensions its recommended grid crosses, each matched by hand to the vendor's counterpart wherever its enumeration contains one; options that only one side proposed are not counted. Agreement runs from 45\% on Paper I, where the two grids diverge most, to 88\% on Paper B, where the vendor independently reproduced the decisive ruling. It also splits by layer: 74\% on the estimation machinery (estimator and functional form, kernel, polynomial order, bandwidth, fixed effects, estimation window, weighting scheme) against 60\% on what is measured and on whom (outcome construct, treatment definition, control group, analysis sample, control set). The specification layer transfers across vendors more readily than the layer that decides what the specification is about. 
The vendor's agents had a licensed Stata installation available inside their sandbox. They used it for five reproductions (Papers B, C, D, G, and J) and two grids (Papers G and H). 
Table~\ref{tab:xvendor_full} places the vendor's ten verdicts beside the main text's. 

\begin{table}[!htbp]
\centering
\caption{The second vendor through the complete framework. Cells: the vendor's grid size, with infeasible cells in parentheses. F's vendor grid covers the paper's secondary leg alone, and its report scored that leg's signs against the primary leg's published direction, which inverts the metric; read against its own leg's direction, 71\% of the vendor's cells carry the published sign, against 68\% of ours for the same leg.}
\label{tab:xvendor_full}
\footnotesize
\setlength{\tabcolsep}{4pt}
\resizebox{\textwidth}{!}{
\begin{tabular}{lcccll}
\hline
 & Cells & Sign \% & Survival \% & & \\
Paper & (infeasible) & (vendor / orig.) & (vendor / orig.) & Vendor's designated inference & Reading \\
\hline
A & 189 (0) & 91 / 100 & 0 / 30 & null-imposed moving-block bootstrap & sign result agrees; survival splits at the designation \\
B & 126 (0) & 94 / 72 & 30 / 9 & assignment-unit-clustered CRV & same: direction holds, significance weak \\
C & 144 (0) & 67 / 95 & 6 / 3 & assignment-level wild cluster bootstrap & same: significance fails \\
D & 216 (0) & 92 / 99 & 0 / 78 & assignment-level RI, ruled uninformative with too few assignment units & splits at the designation \\
E & 192 (128) & 0 / 22 & 0 / 0 & assignment-level wild cluster bootstrap-t & same: no support \\
F & 160 (0) & 29 / 68 & 0 / 20 & two-way CRV1 & secondary leg only (primary leg 67); see caption \\
G & 216 (0) & 88 / 95 & 6 / 17 & assignment-level wild cluster bootstrap & same: marginal \\
H & 120 (0) & 100 / 77 & 38 / 0 & assignment-cell wild cluster bootstrap & splits at the designation \\
I & 144 (0) & 50 / 64 & 46 / 10 & one-way unit-level CRV, heterogeneity-robust DiD estimator & trend control not enumerated \\
J & 240 (120) & 48 / 93 & 0 / 47 & unit-level wild cluster bootstrap-t & same in substance: effect indistinguishable from zero \\
\hline
\end{tabular}}
\end{table}

Six of the ten main verdicts agree. On B, C, E, F, G, and J the two runs reach the same reading of the published conclusion by independent routes. For J, the vendor's analog of the published specification returns an estimate indistinguishable from zero (bootstrap $p = 0.81$), which is where the main text also lands, although it gets there through the placebo diagnostic rather than through the point estimate. 

The four remaining disagreements have two distinct causes. Three of them, on Papers A, D, and H, are not disagreements about any estimate. They are disagreements about which inference method should be the main effect, the thing the validation agent decides at Step~2 before a single coefficient exists. On A the two runs agree that essentially every cell carries the published sign and differ only on whether the curve is significant, because the vendor imposes a stricter null-imposed bootstrap. On D the vendor holds that the number of assignment units is too small to support an informative $p$-value at all. On H the vendor designates a wild cluster bootstrap where the main text designates randomization inference. These are the same three designations the third model judges below.

The fourth disagreement, Paper I, comes from the fork space instead. The second model's enumeration never proposes the group-by-time trend control that the main text's grid imposes in every cell, and its map reads 46\% survival where the main text's is 10\%. This is the practical argument for pooling fork proposals from several models into a single verified grid. Also we note that having Stata available does not change the main qualitative result. 

\paragraph{A third-model judgment of the inference designation.}
The fork validation designates the inference method before any estimate exists, and this designation turned out to be the largest source of cross-model disagreement. We therefore re-ran Step 2 alone with a third model, \texttt{gpt-6-astra} (two model generations after \texttt{gpt-5.6-sol}, and the newest generation available at the time of writing), on four papers. Papers A, D, and H were selected because the two earlier audits had designated different headline inference methods.

The four judgments sort into two cases. On Paper B, where the decisive fork is the clustering level, the third model reproduces the designation the other two reached independently. Clustering below the treatment unit is invalid, assignment-level clustering binds, and the second identification strategy stays outside the primary grid as corroboration. On the three papers where defensible inference schemes compete, the designations split, and the third model is the most conservative of the three. On H it adopts the second vendor's wild cluster bootstrap, matching the cluster variable and replication count, and rates the exact randomization test questionable on estimand-isolation grounds. On D it holds that no supplied method yields an informative headline $p$-value with so few assignment units, the stricter branch of the second vendor's own reading, and rejects the in-time placebo route as a falsification diagnostic rather than an inference method. On A it declines to certify any headline $p$-value from the enumerated forks, rating the displaced-cutoff permutation invalid and the block bootstrap questionable, and states the conditions a valid procedure would have to meet. Such cross-model analysis shows the difficulty in evaluating assumptions in empirical research. Here we refrain from discussing the design choice of each individual papers and argue that,  AI models indeed 
find important robustness cues in empirical research.

\end{document}